\documentclass[nohyperref]{article}
\usepackage[accepted]{icml2022}
\icmltitlerunning{Structure-Aware Graph Transformer}

\usepackage{microtype}
\usepackage{graphicx}
\usepackage{subcaption}     % for subfigures
\usepackage{booktabs}    % for professional tables
\usepackage{comment}     % for block comments
\usepackage{dsfont}      % for Reals    
\usepackage{multirow}    % multi row in tables
\usepackage{stmaryrd}  % to have short up/down arrows in tables
\usepackage{paralist}  % in line lists
\usepackage{enumitem}   % to make compact lists
\usepackage{censor}     % to create a censored item

\usepackage[hidelinks,pagebackref=True]{hyperref} 
\renewcommand*\backref[1]{\ifx#1\relax \else (Cited on #1) \fi}

\newcommand{\leo}[1]{{\color{black}#1}}

\usepackage{amsmath}
\usepackage{amssymb}
\usepackage{mathtools}
\usepackage{amsthm}

\usepackage{pgf, tikz}
\usepackage{pgfplots}
\usepackage{pgfplotstable}

\usetikzlibrary{arrows, automata}
\pgfplotsset{compat=1.15}
\pgfplotsset{every axis/.append style={font=\Large}}
\pgfplotsset{every tick label/.append style={font=\Large}}

\usetikzlibrary{arrows}
\usetikzlibrary{calc}
\usetikzlibrary{chains}
\usetikzlibrary{fit}
\usetikzlibrary{matrix}
\usetikzlibrary{positioning}
\usetikzlibrary{scopes}
\usetikzlibrary{external}
\usetikzlibrary{shadows}
\usetikzlibrary{shadows.blur}

\pgfplotsset{compat=1.16}
\usepgfplotslibrary{colorbrewer}
\usepgfplotslibrary{groupplots}
\usepgfplotslibrary{colormaps}

\pgfplotsset{
  compat                 = 1.16,  % Okay for Overleaf, I guess...
  filter discard warning = false, % Suppress warnings for filtered plots
  discard if not/.style 2 args={
    x filter/.append code={
      \edef\tempa{\thisrow{#1}}
      \edef\tempb{#2}
      \ifx\tempa%
        \tempb%
      \else%
        
      \fi
    }
  },
}

\pgfplotsset{%
  every non boxed x axis/.append style={x axis line style=-},
  every non boxed y axis/.append style={y axis line style=-}
}

\usepgfplotslibrary{colorbrewer}
\usepgfplotslibrary{groupplots}
\usetikzlibrary{positioning}
\usetikzlibrary{chains}

\definecolor{538red}{RGB}{129, 15, 124}
\definecolor{emerald}{RGB}{15,132,101}
\definecolor{lightgray}{RGB}{211,211,211}
\definecolor{coral}{RGB}{200,42,42}
\definecolor{deepblue}{RGB}{11,83,148}
\definecolor{brightblue}{RGB}{56,122,243}

\usepackage[capitalize,noabbrev]{cleveref}

\theoremstyle{plain}
\newtheorem{theorem}{Theorem}

\theoremstyle{definition}

\theoremstyle{remark}

\newcommand{\graph}     {\ensuremath{G}}
\newcommand{\vertices}  {\ensuremath{V}}
\newcommand{\edges}     {\ensuremath{E}}
\newcommand{\X}         {\ensuremath{\mathbf{X}}}
\newcommand{\Q}         {\ensuremath{\mathbf{Q}}}
\newcommand{\K}         {\ensuremath{\mathbf{K}}}
\newcommand{\V}         {\ensuremath{\mathbf{V}}}
\newcommand{\Wq}         {\ensuremath{\mathbf{W_Q}}}
\newcommand{\Wk}         {\ensuremath{\mathbf{W_K}}}
\newcommand{\Wv}         {\ensuremath{\mathbf{W_V}}}
\def\Xcal{{\mathcal X}}
\def\Hcal{{\mathcal H}}
\def\H{{\mathbf H}}
\def\Ncal{{\mathcal N}}
\newcommand{\ie}{\textit{i.e.\ }}
\newcommand{\eg}{\textit{e.g.\ }}

\newcommand{\real}{\ensuremath{\mathds{R}}}
\newcommand{\AGGREGATE} {\texttt{AGGREGATE}}
\newcommand{\COMBINE}   {\texttt{COMBINE}}
\newcommand{\READOUT}   {\texttt{READOUT}}
\newcommand{\softmax}   {\ensuremath{\mathrm{softmax}}}
\usepackage[textsize=tiny]{todonotes}

\begin{document}

\twocolumn[
\icmltitle{Structure-Aware Transformer for Graph Representation Learning}
\setitemize{noitemsep,topsep=0pt,parsep=0pt,partopsep=0pt} % compact list
\icmlsetsymbol{equal}{*}

\begin{icmlauthorlist}
\icmlauthor{Dexiong Chen}{equal,eth,sib}
\icmlauthor{Leslie O'Bray}{equal,eth,sib}
\icmlauthor{Karsten Borgwardt}{eth,sib}
\end{icmlauthorlist}

\icmlaffiliation{eth}{Department of Biosystems Science and Engineering, ETH Zürich, Switzerland}
\icmlaffiliation{sib}{SIB Swiss Institute of Bioinformatics, Switzerland}

\icmlcorrespondingauthor{Dexiong Chen}{dexiong.chen@bsse.ethz.ch}
\icmlcorrespondingauthor{Leslie O'Bray}{leslie.obray@bsse.ethz.ch}

\icmlkeywords{graph transformer, graph neural network, graph kernel}
\vskip 0.3in
]
\printAffiliationsAndNotice{\icmlEqualContribution} % otherwise use the standard text.

\begin{abstract}
    The Transformer architecture has gained growing attention in graph representation learning recently, as it naturally overcomes several limitations of graph neural networks (GNNs) by avoiding their strict structural inductive biases and instead only encoding the graph structure via positional encoding. Here, we show that the node representations generated by the Transformer with positional encoding do not necessarily capture structural similarity between them. To address this issue, we propose the Structure-Aware Transformer, a class of simple and flexible graph Transformers built upon a new self-attention mechanism. This new self-attention incorporates structural information into the original self-attention by extracting a subgraph representation rooted at each node before computing the attention. We propose several methods for automatically generating the subgraph representation and show theoretically that the resulting representations are at least as expressive as the subgraph representations. Empirically, our method achieves state-of-the-art performance on five graph prediction benchmarks. Our structure-aware framework can leverage any existing GNN to extract the subgraph representation, and we show that it systematically improves performance relative to the base GNN model, successfully combining the advantages of GNNs and Transformers. \leo{Our code is available at \url{https://github.com/BorgwardtLab/SAT}.}
\end{abstract}

\section{Introduction}\label{sec:introduction}

Graph neural networks (GNNs) have been established as powerful and flexible tools for graph representation learning, with successful applications in drug discovery~\citep{gaudelet2021utilizing}, protein design~\citep{ingraham2019generative}, social network analysis~\citep{fan2019graph}, and so on. A large class of GNNs build multilayer models, where each layer operates on the previous layer to generate new representations using a message-passing mechanism~\citep{gilmer2017neural} to aggregate local neighborhood information. 

While many different message-passing strategies have been proposed, some critical limitations have been uncovered in this class of GNNs. These include the limited expressiveness of GNNs~\citep{xu2018how,morris2019weisfeiler}, as well as known problems such as over-smoothing~\citep{Li2018, li2019deepgcns, chen2020measuring, oono2020graph} and over-squashing~\citep{alon2021on}. 
%While stemming from a similar property of graphs---the exponential receptive size of a node as the distance increases from it---over-smoothing and over-squashing are separate phenomena. 
Over-smoothing manifests as all node representations converging to a constant after sufficiently many layers, while over-squashing occurs when messages from distant nodes are not effectively propagated through certain ``bottlenecks'' in a graph, since too many messages get compressed into a single fixed-length vector. Designing new architectures beyond neighborhood aggregation is thus essential to solve these problems.

\begin{figure}[t]
    
     \resizebox{0.95\columnwidth}{!}{
     \begin{subfigure}[b]{0.45\columnwidth}
         \centering

         \begin{tikzpicture}
      \tikzset{
        every node/.style = {
          shape        = circle,
          minimum size = 6pt,
          inner sep    = 0pt,
          fill         = black,
          draw         = black,
          font         = \small,
        },
      }
      \begin{scope}[scale=0.5]
        
        \node[fill=deepblue, draw=deepblue] (N2) at (1, -0) {};
        \node[fill=deepblue, draw=deepblue, label={$u$}] (N3) at (-1, -0) {};
        \node[fill=coral, draw=coral] (N4) at (-2, -1) {};
        \node[fill=coral, draw=coral] (N5) at (2, 1) {};
        \node[fill=coral, draw=coral] (N6) at (2, -1) {};
        \node[fill=coral, draw=coral] (N7) at (-2, 1) {};
        \node[fill=white, draw=white] (Title) at (0, 2) {$G_1$};
        
        \draw (N2) -- (N3);
        \draw (N3) -- (N7);
        \draw (N3) -- (N4);
        
        \draw (N2) -- (N5);
        \draw (N2) -- (N6);
        %\draw (N7) -- (N4);
        
      \end{scope}

 \end{tikzpicture}
    
     \end{subfigure}
     \hfill
     \begin{subfigure}[b]{0.45\columnwidth}
         \centering
         \begin{tikzpicture}
      \tikzset{
        every node/.style = {
          shape        = circle,
          minimum size = 6pt,
          inner sep    = 0pt,
          fill         = black,
          draw         = black,
          font         = \small,
        },
      }
      \begin{scope}[scale=0.5]
        \node[fill=deepblue, draw=deepblue] (N2) at (1, -0) {};
        \node[fill=deepblue, draw=deepblue,label={$v$}] (N3) at (-1, -0) {};
        \node[fill=coral, draw=coral] (N4) at (-2, -1) {};
        \node[fill=coral, draw=coral] (N5) at (2, 1) {};
        \node[fill=coral, draw=coral] (N6) at (2, -1) {};
        \node[fill=coral, draw=coral] (N7) at (-2, 1) {};
        \node[fill=white, draw=white] (Title) at (0,2) {$G_2$};
        
        \draw (N2) -- (N3);
        \draw (N3) -- (N7);
        \draw (N3) -- (N4);
        
        \draw (N2) -- (N5);
        \draw (N2) -- (N6);
        \draw (N7) -- (N4);
        
      \end{scope}

    \end{tikzpicture}

     \end{subfigure}
     \hfill
     
     }
        \caption{\emph{Position-aware} vs. \emph{structure-aware}: Using a positional encoding based on shortest paths in $G_1$ and $G_2$ respectively (assuming all edges have equal weight), node $u$ and $v$ would receive identical encodings since their shortest paths to all other nodes are the same in both graphs. However, their structures are different, with $v$ forming a triangle with its red neighbors.\vspace{-0.4cm} }
        \label{fig:toy_example}
\end{figure}
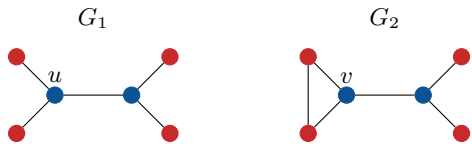

Transformers~\citep{vaswani2017attention}, which have proved to be successful in natural language understanding~\citep{vaswani2017attention}, computer vision~\citep{dosovitskiy2020image}, and biological sequence modeling~\citep{rives2021biological}, offer the potential to address these issues. Rather than only aggregating local neighborhood information in the message-passing mechanism, the Transformer architecture is able to capture interaction information between any node pair via a single self-attention layer. 
Moreover, in contrast to GNNs, the Transformer avoids introducing any structural inductive bias at intermediate layers, addressing the expressivity limitation of GNNs. Instead, it encodes structural or positional information about nodes only into input node features, albeit limiting how much information it can learn from the graph structure. Integrating information about the graph structure into the Transformer architecture has thus gained growing attention in the graph representation learning field. However, most existing approaches only encode positional relationships between nodes, rather than explicitly encoding the structural relationships. As a result, they may not identify structural similarities between nodes and could fail to model the \emph{structural interaction} between nodes (see Figure~\ref{fig:toy_example}). This could explain why their performance was dominated by sparse GNNs in several tasks~\citep{dwivedi2022graph}.

\paragraph{Contribution}

In this work, we address the critical question of how to encode structural information into a Transformer architecture. Our principal contribution is to introduce a flexible \emph{structure-aware} self-attention mechanism that explicitly considers the graph structure and thus captures structural interaction between nodes. The resulting class of Transformers, which we call the Structure-Aware Transformer (SAT), can provide structure-aware representations of graphs, in contrast to most existing position-aware Transformers for graph-structured data. Specifically: 
\begin{compactitem}
    \item We reformulate the self-attention mechanism in \citet{vaswani2017attention} as a kernel smoother and
        extend the original exponential kernel on node features to also account for local structures, by extracting a subgraph representation centered around each node. 
    \item  We propose several methods for automatically generating the subgraph representations, enabling the resulting kernel smoother to simultaneously capture structural and attributed similarities between nodes. The resulting representations are theoretically guaranteed to be at least as expressive as the subgraph representations. 
    \item We demonstrate the effectiveness of SAT models on five graph and node property prediction benchmarks by showing it achieves better performance than state-of-the-art GNNs and Transformers. Furthermore, we show how SAT can easily leverage any GNN to compute the \leo{node representations which incorporate subgraph information} and outperform the base GNN, making it an effortless enhancer of any existing GNN.
    \item Finally, we show that we can attribute the performance gains to the structure-aware aspect of our architecture, and showcase how SAT is more interpretable than the classic Transformer with an absolute encoding.
\end{compactitem}

We will present the related work and relevant background in Sections~\ref{sec:Related Work} and \ref{sec:background} before presenting our method in Section~\ref{sec:Method} and our experimental findings in Section~\ref{sec:experiments}.

%%%%%%%%%%%%%%%%%%%%%%%%%%%%%%%%%%%%%%%%%%%%%%%%%%%%%%%%%%%%%%%
% Toy example figure
%%%%%%%%%%%%%%%%%%%%%%%%%%%%%%%%%%%%%%%%%%%%%%%%%%%%%%%%%%%%%%%
%%%%%%%%%%%%%%%%%%%%%%%%%%%%%%%%%%%%%%%%%%%%%%%%%%%%%%%%%%%%%%%

\section{Related Work}\label{sec:Related Work}
We present here the work most related to ours, namely the work stemming from message passing GNNs, positional representations on graphs, and graph Transformers. 

\paragraph{Message passing graph neural networks }

Message passing graph neural networks have recently been one of the leading methods for graph representation learning. An early seminal example is the GCN~\citep{Kipf2017}, which was based on performing convolutions on the graph. \citet{gilmer2017neural} reformulated the early GNNs into a framework of message passing GNNs, which has since then become the predominant framework of GNNs in use today, \leo{with extensive examples \citep{hamilton2017inductive, xu2018how, corso2020principal, hu2020pretraining, velickovic2018graph, li2020deepergcn, Yang2022}.} However, as mentioned above, they suffer from problems of limited expressiveness, over-smoothing, and over-squashing.

\paragraph{Absolute encoding }

Because of the limited expressiveness of GNNs, there has been some recent research into the use of \emph{absolute encoding}~\citep{shaw2018self}, which consists of adding or concatenating positional or structural representations to the input node features. While it is often called an \emph{absolute positional encoding}, we refer to it more generally as an \emph{absolute encoding} to include both positional and structural encoding, which are both important in graph modeling. Absolute encoding primarily considers position or location relationships between nodes. Examples of position-based methods include the Laplacian positional encoding~\citep{dwivedi2021generalization, kreuzer2021rethinking}, Weisfeiler--Lehman-based positional encoding~\citep{zhang2020graph}, and random walk positional encoding (RWPE)~\citep{li2020distance,dwivedi2022graph}, while distance-based methods include distances to a predefined set of nodes~\citep{you2019position} and shortest path distances between pairs of nodes \citep{zhang2020graph, li2020distance}. \citet{dwivedi2022graph} extend these ideas by using a trainable absolute encoding.

\paragraph{Graph Transformers }

While the absolute encoding methods listed above can be used with message passing GNNs, they also play a crucial role in the (graph) Transformer architecture. Graph Transformer~\citep{dwivedi2021generalization} provided an early example of how to generalize the Transformer architecture to graphs, using Laplacian eigenvectors as an absolute encoding and computing attention on the immediate neighborhood of each node, rather than on the full graph. SAN~\citep{kreuzer2021rethinking} also used the Laplacian eigenvectors for computing an absolute encoding, but computed attention on the full graph, while distinguishing between true and created edges. Many graph Transformer methods also use a \emph{relative encoding}~\citep{shaw2018self} in addition to absolute encoding. This strategy incorporates representations of the relative position or distances between nodes on the graph directly into the self-attention mechanism, as opposed to the absolute encoding which is only applied once to the input node features.  \citet{mialon2021graphit} propose a relative encoding by means of kernels on graphs to bias the self-attention calculation, which is then able to incorporate positional information into Transformers via the choice of kernel function. Other recent work seeks to incorporate structural information into the graph Transformer, for example by encoding some carefully selected graph theoretic properties such as centrality measures and shortest path distances as positional representations~\citep{ying2021do} or by using GNNs to integrate the graph structure~\citep{Rong2020, jain2021representing,mialon2021graphit, Shi2021}. 

In this work, we combine the best of both worlds from message passing GNNs and from the Transformer architecture. We incorporate both an absolute as well as a novel relative encoding that explicitly incorporates the graph structure, thereby designing a Transformer architecture that takes both local and global information into account.

\section{Background}\label{sec:background}

In the following, we refer to a graph as $\graph = (\vertices, \edges, 
\X)$, where the node attributes for node $u\in V$ is denoted by $x_u\in\Xcal\subset\real^d$ and the node attributes for all nodes are stored in $\X \in \real^{n \times d}$ for a graph with $n$ nodes.

\subsection{Transformers on Graphs}\label{sec:Transformers}
While GNNs use the graph structure explicitly, Transformers remove that explicit structure, and instead infer relations between nodes by leveraging the node attributes. In this sense, the Transformer~\citep{vaswani2017attention} ignores the graph structure and rather considers the graph as a (multi\nobreakdash-) set of nodes, and uses the self-attention mechanism to infer the similarity between nodes. The Transformer itself is composed of two main blocks: a self-attention module followed by a feed-forward neural network. In the self-attention module, the input node features $\X$ are first projected to query ($\Q$), key ($\K$) and value ($\V$) matrices through a linear projection such that $\Q=\X \Wq$, $\K=\X\Wk$ and $\V=\X \Wv$ respectively. We can compute the self-attention via
    \begin{equation}\label{eq:Attention}
       \mathrm{Attn}(\X):=\softmax(\frac{\Q\K^T}{\sqrt{d_{out}}})\V\in\real^{n\times d_{out}},
    \end{equation}
where $d_{out}$ refers to the dimension of $\Q$, and $\Wq, \Wk, \Wv$ are trainable parameters. It is common to use \emph{multi-head} attention, which concatenates multiple instances of Eq.~\eqref{eq:Attention} and has shown to be effective in practice~\citep{vaswani2017attention}. Then, the output of the self-attention is followed by a skip-connection and a feed-forward network (FFN), which jointly compose a Transformer layer, as shown below:
\begin{equation}\label{eq:skip+FFN}
\begin{aligned}
    \X' &= \X + \mathrm{Attn}(\X), \\
    \X'' &=\mathrm{FFN}(\X'):=\text{ReLU}(\X' W_1)W_2.
\end{aligned}
\end{equation}
Multiple layers can be stacked to form a Transformer model, which ultimately provides node-level representations of the graph. As the self-attention is equivariant to permutations of the input nodes, the Transformer will always generate the same representations for nodes with the same attributes regardless of their locations and surrounding structures in the graph. It is thus necessary to incorporate such information into the Transformer, generally via absolute encoding.

\paragraph{Absolute encoding }

Absolute encoding refers to adding or concatenating the positional or structural representations of the graph to the input node features before the main Transformer model, such as the Laplacian positional encoding~\citep{dwivedi2021generalization} or RWPE~\citep{dwivedi2022graph}. The main shortcoming of these encoding methods is that they generally do not provide a measure of the structural similarity between nodes and their neighborhoods.

%%%%%%%%%%%%%%%%%%%%%%%%% Overview Figure %%%%%%%%%%%%%%%%%%%%%%%%%%%
\input{figures/overview}
%%%%%%%%%%%%%%%%%%%%%%%%% Overview Figure %%%%%%%%%%%%%%%%%%%%%%%%%%%

\paragraph{Self-attention as kernel smoothing }

As noticed by~\citet{mialon2021graphit}, the self-attention in Eq.~\eqref{eq:Attention} can be rewritten as a kernel smoother
\begin{equation}
    \mathrm{Attn}(x_v)=\sum_{u\in V} \frac{\kappa_{\exp}(x_v, x_u)}{\sum_{w\in V} \kappa_{\exp}(x_v,x_w)}f(x_u),~\forall v\in V,
\end{equation}
where $f(x)=\Wv x$ is the linear value function and $\kappa_{\exp}$ is a (non-symmetric) exponential kernel on $\real^d\times\real^d$ parameterized by $\Wq$ and $\Wk$:
\begin{equation}
    \kappa_{\exp}(x, x'):=\exp\left(\langle\Wq x,\Wk x'\rangle/\sqrt{d_{out}}\right),
\end{equation}
where $\langle \cdot,\cdot \rangle$ is the dot product on $\real^d$. With this form, \citet{mialon2021graphit} propose a relative positional encoding strategy via the product of this kernel and a diffusion kernel on the graph, which consequently captures the positional similarity between nodes. However, this method is only position-aware, in contrast to our structure-aware encoding that will be presented in Section~\ref{sec:Method}.

\section{Structure-Aware Transformer}\label{sec:Method}
In this section, we will describe how to encode the graph structure into the self-attention mechanism and provide a class of Transformer models based on this framework.

\subsection{Structure-Aware Self-Attention}\label{subsec:sw-attn}
As presented above, self-attention in the Transformer can be rewritten as a kernel smoother where the kernel is a trainable exponential kernel defined on node features, and which only captures attributed similarity between a pair of nodes. The problem with this kernel smoother is that it cannot filter out nodes that are structurally different from the node of interest when they have the same or similar node features. In order to also incorporate the structural similarity between nodes, we consider a more generalized kernel that additionally accounts for the local substructures around each node. By introducing a set of subgraphs centered at each node, we define our structure-aware attention as:
\begin{equation}\label{eq:sw-attn}
    \text{SA-attn}(v):=\sum_{u\in V} \frac{\kappa_{\text{graph}}(S_{G}(v), S_{G}(u))}{\sum_{w\in V} \kappa_{\text{graph}}(S_{G}(v), S_{G}(w))}f(x_u),
\end{equation}
where $S_{G}(v)$ denotes a subgraph in $G$ centered at a node $v$ associated with node features $\X$ and $\kappa_{\text{graph}}$ can be any kernel that compares a pair of subgraphs. This new self-attention function not only takes the attributed similarity into account but also the structural similarity between subgraphs. It thus generates more expressive node representations than the original self-attention, as we will show in Section~\ref{sec:expressiveness}. Moreover, this self-attention is no longer equivariant to any permutation of nodes but only to nodes whose features and subgraphs coincide, which is a desirable property.

In the rest of the paper, we will consider the following form of $\kappa_{\text{graph}}$ that already includes a large class of expressive and computationally tractable models:
\begin{equation}
    \kappa_{\text{graph}}(S_{G}(v), S_{G}(u))=\kappa_{\exp}(\varphi(v,G), \varphi(u,G)),
\end{equation}
where $\varphi(u,G)$ is a \emph{structure extractor} that extracts vector representations of some subgraph centered at $u$ with node features $\X$. We provide several alternatives of the structure extractor below. It is worth noting that our structure-aware self-attention is flexible enough to be combined with any model that generates representations of subgraphs, including GNNs and (differentiable) graph kernels. For notational simplicity, we assume there are no edge attributes, but our method can easily incorporate edge attributes as long as the structure extractor can accommodate them. \leo{The edge attributes are consequently not considered in the self-attention computation, but are incorporated into the structure-aware node representations. In the structure extractors presented in this paper, this means that edge attributes were included whenever the base GNN was able to handle edge attributes.}

\paragraph{$k$-subtree GNN extractor}

A straightforward way to extract local structural information at node $u$ is to apply any existing GNN model to the input graph with node features $\X$ and take the output node representation at $u$ as the subgraph representation at $u$. More formally, if we denote by $\text{GNN}_{G}^{(k)}$ an arbitrary GNN model with $k$ layers applied to $G$ with node features $\X$, then
\begin{equation}
    \varphi(u, G)=\text{GNN}^{(k)}_{G}(u).
\end{equation}
This extractor is able to represent the $k$-subtree structure rooted at $u$~\citep{xu2018how}. While this class of structure extractors is fast to compute and can flexibly leverage any existing GNN, they cannot be more expressive than the Weisfeiler--Lehman test due to the expressiveness limitation of message passing GNNs~\citep{xu2018how}. In practice, a small value of $k$ already leads to good performance, while not suffering from over-smoothing or over-squashing.

\paragraph{$k$-subgraph GNN extractor}
A more expressive extractor is to use a GNN to directly compute the representation of the entire $k$-hop subgraph centered at $u$ rather than just the node representation $u$. \leo{Recent work has explored the idea of using subgraphs rather than subtrees around a node in GNNs, with positive experimental results \citep{zhang2021nested, wijesinghe2022a}, as well as being strictly more powerful than the 1-WL test \citep{zhang2021nested}. We follow the same setup as is done in \citet{zhang2021nested}, and adapt our GNN extractor to utilize the entire $k$-hop subgraph.}
The $k$-subgraph GNN extractor aggregates the updated node representations of all nodes within the $k$-hop neighborhood using a pooling function such as summation. Formally, if we denote by $\Ncal_k(u)$ the $k$-hop neighborhood of node $u$ including itself, the representation of a node $u$ is:
\begin{equation}
    \varphi(u, G)=\sum_{v\in\Ncal_k(u)}\text{GNN}^{(k)}_{G}(v).
\end{equation}

We observe that prior to the pooling function, the $k$-subgraph GNN extractor is equivalent to using the $k$-subtree GNN extractor within each $k$-hop subgraph. So as to capture the attributed similarity as well as structural similarity, we augment the node representation from $k$-subgraph GNN extractor with the original node features via concatenation. While this extractor provides more expressive subgraph representations than the $k$-subtree extractor, it requires enumerating all $k$-hop subgraphs, and consequently does not scale as well as the $k$-subtree extractor to large datasets.

\paragraph{Other structure extractors}

Finally, we present a list of other potential structure extractors for different purposes. One possible choice is to directly learn a number of ``hidden graphs'' as the ``anchor subgraphs'' to represent subgraphs for better model interpretability, by using the concepts introduced in~\citet{nikolentzos2020random}. While \citet{nikolentzos2020random} obtain a vector representation of the input graph by counting the number of matching walks between the whole graph and each of the hidden graphs, one could extend this to the node level by comparing the hidden graphs to the $k$-hop subgraph centered around each node. The adjacency matrix of the hidden graphs is a trainable parameter in the network, thereby enabling end-to-end training to identify which subgraph structures are predictive. Then, for a trained model, visualizing the learned hidden graphs provides useful insights about the structural motifs in the dataset.

Furthermore, more domain-specific GNNs could also be used to extract potentially more expressive subgraph representations. For instance, \citet{bodnar2021weisfeiler} recently proposed a new kind of message passing scheme operating on regular cell complexes which benefits from provably stronger expressivity for molecules. Our self-attention mechanism can fully benefit from the development of more domain-specific and expressive GNNs.

Finally, another possible structure extractor is to use a non-parametric graph kernel (\eg a Weisfeiler-Lehman graph kernel) on the $k$-hop subgraphs centered around each node. This provides a flexible way to combine graph kernels and deep learning, which might offer new theoretical insights into the link between the self-attention and kernel methods.

\subsection{Structure-Aware Transformer}\label{sec:pooling}
Having defined our structure-aware self-attention function, the other components of the Structure-Aware Transformer follow the Transformer architecture as described in Section~\ref{sec:Transformers}; see Figure~\ref{fig:Overview} for a visual overview. Specifically, the self-attention function is followed by a skip-connection, a FFN and two normalization layers before and after the FFN. In addition, we also include the degree factor in the skip-connection, which was found useful for reducing the overwhelming influence of highly connected graph components~\citep{mialon2021graphit}, i.e.,
\begin{equation}
    x_v'=x_v+1/\sqrt{d_v} \,\text{SA-attn}(v),
\end{equation}
where $d_v$ denotes the degree of node $v$. After a Transformer layer, we obtain a new graph with the same structure but different node features $G'=(V,E,\X')$, where $\X'$ corresponds to the output of the Transformer layer.

Finally, for graph property prediction, there are various ways to aggregate node-level representations into a graph representation, such as by taking the average or sum. Alternatively, one can use the embedding of a virtual [CLS] node \cite{jain2021representing} that is attached to the input graph without any connectivity to other nodes. We compare these approaches in Section~\ref{sec:experiments}.

\subsection{Combination with Absolute Encoding}

While the self-attention in Eq.~\eqref{eq:sw-attn} is structure-aware, most absolute encoding techniques are only position-aware and could therefore provide complementary information. Indeed, we find that the combination leads to further performance improvements, which we show in Section~\ref{sec:experiments}. We choose to use the RWPE~\citep{dwivedi2022graph}, though any other absolute positional representations, including learnable ones, can also be used. 

We further argue that only using absolute positional encoding with the Transformer would exhibit a too relaxed structural inductive bias which is not guaranteed to generate similar node representations even if two nodes have similar local structures. This is due to the fact that distance or Laplacian-based positional representations generally serve as structural or positional signatures but do not provide a measure of structural similarity between nodes, especially in the inductive case where two nodes are from different graphs. This is also empirically affirmed in Section~\ref{sec:experiments} by their relatively worse performance without using our structural encoding. In contrast, the subgraph representations used in the structure-aware attention can be tailored to measure the structural similarity between nodes, and thus generate similar node-level representations if they possess similar attributes and surrounding structures. We can formally state this in the following theorem:
\begin{theorem}
    Assume that $f$ is a Lipschitz mapping with the Lipschitz constant denoted by $\text{Lip}(f)$ and the structure extractor $\varphi$ is bounded by a constant $C_{\varphi}$ on the space of subgraphs. For any pair of nodes $v$ and $v'$ in two graphs $G=(V,E,\X)$ and $G'=(V',E',\X')$ with the same number of nodes $|V|=|V'|$, the distance between their representations after the structure-aware attention is bounded by:
    \begin{equation}\label{eq:invariance}
        \begin{split}
         \|\text{SA-attn}(v)-\text{SA-attn}(v')\|\leq C_1 [\|h_v - h_{v'}'\| \\ +D(\H, \H')] +C_2 D(\X,\X'),
        \end{split}
    \end{equation}
    where $C_1, C_2>0$ are constants depending on $|V|$, $\text{Lip}(f)$, $C_{\varphi}$ and spectral norms of the parameters in SA-attn, whose expressions are given in the Appendix, and $h_w:=\varphi(w,G)$ denotes the subgraph representation at node $w$ for any $w\in V$ and $h_{w'}':=\varphi(w',G')$ similarly, and $\H=(h_w)_{w\in V}$ and $\H'=(h_{w'}')_{w'\in V'}$ denote the multiset of subgraph representations in $G$ and $G'$ respectively. Denoting by $\Pi(V,V')$ the set of permutations from $V$ to $V'$, $D$ is an optimal matching metric between two multisets of representations with the same cardinality, defined as
    \begin{equation*}
        D(\X,\X'):=\inf_{\pi\in\Pi(V,V')}\sup_{w\in V} \|x_w-x_{\pi(w)}'\|.
    \end{equation*}
    %
    % $C_1$ and $C_2$ are constants given by:
    % \begin{equation*}
    %     C_1=\sqrt{\frac{2}{d_{out}}}n\text{Lip}(f) C_{\varphi} \|\Wq\|_{\infty} \|\Wk\|_{\infty}, ~~~
    %     C_2=\text{Lip}(f).
    % \end{equation*}
\end{theorem}

The proof is provided in the Appendix. The metric $D$ is an optimal matching metric between two multisets which measures how different they are. This theorem shows that two node representations from the $\text{SA-attn}$ are similar if the graphs that they belong to have similar multisets of node features and subgraph representations overall, and at the same time, the subgraph representations at these two nodes are similar. In particular, if two nodes belong to the same graph, \ie $G=G'$, then the second and last terms on the right side of Eq.~\eqref{eq:invariance} are equal to zero and the distance between their representations is thus constrained by the distance between their corresponding subgraph representations. However, for Transformers with absolute positional encoding, the distance between two node representations is not constrained by their structural similarity, as the distance between two positional representations does not necessarily characterize how structurally similar two nodes are. Despite stronger inductive biases, we will show that our model is still sufficiently expressive in the next section.

%%%%%%%%%%%%%%%%%%%%%%%%%%%%%%%%%%%%%%%%%%%%%%%%%%%%%%%%%%%%%%%%%%%%%%%%
% SOTA Table
%%%%%%%%%%%%%%%%%%%%%%%%%%%%%%%%%%%%%%%%%%%%%%%%%%%%%%%%%%%%%%%%%%%%%%%%

%%%%%%%%%%%%%%%%%%%%%%%%%%%%%%%%%%%%%%%%%%%%%%%%%%%%%%%%%%%%%%%%%%%%%%%%
% Comparison to SOTA Transformers
%%%%%%%%%%%%%%%%%%%%%%%%%%%%%%%%%%%%%%%%%%%%%%%%%%%%%%%%%%%%%%%%%%%%%%%%

\begin{table}[tbp]
  \caption{Comparison of SAT to SOTA methods on graph regression and classification tasks. ZINC results use edge weights where applicable, otherwise without edge weights. $^\star$ indicates we obtained the results ourselves by adapting the code provided by the original paper. \leo{ $\shortuparrow$ means that higher is better for the performance metric; $\shortdownarrow$ indicates lower is better.}}
  \label{tab:comparison_to_SOTA}
  \centering
    \begin{small}
    \setlength{\tabcolsep}{0.50em}
      \begin{sc}
      
        \resizebox{.48\textwidth}{!}{%
      
            \begin{tabular}{lcccc}
            \toprule
            {}&\multicolumn{1}{c}{\textbf{ZINC} $\shortdownarrow$}&\multicolumn{1}{c}{\textbf{CLUSTER} $\shortuparrow$} &\multicolumn{1}{c}{\textbf{PATTERN} $\shortuparrow$} \vspace{0.5em} \\
            \# graphs & 12,000 & 12,000 & 14,000 \\
            Avg. \# nodes & 23.2 & 117.2 & 118.9\\
            Avg. \# edges & 49.8 & 4,303.9 & 6,098.9 \\
            Metric & MAE & Accuracy & Accuracy\\
            \midrule
            GIN & 0.387$\pm$0.015 & 64.716$\pm$1.553 & 85.590$\pm$0.011  \\
            GAT & 0.384$\pm$0.007 & 70.587$\pm$0.447 &  78.271$\pm$0.186 \\
            PNA & 0.188$\pm$0.004 & 67.077$\pm$0.977$^\star$ & 86.567$\pm$0.075 \\
            \midrule
            Transformer+RWPE & 0.310$\pm$0.005 & 29.622$\pm$0.176 & 86.183$\pm$0.019 \\
            Graph Transformer & 0.226$\pm$0.014 & 73.169$\pm$0.622 &  84.808$\pm$0.068 \\
            SAN & 0.139$\pm$0.006 & 76.691$\pm$0.650 & 86.581$\pm$0.037 \\
            Graphormer & 0.122$\pm$0.006 & -- & -- \\
            \midrule
            k-subtree SAT & 0.102$\pm$0.005 & 77.751$\pm$0.121 & \textbf{86.865$\pm$0.043} \\
            k-subgraph SAT & \textbf{0.094$\pm$0.008} & \textbf{77.856$\pm$0.104} & 86.848$\pm$0.037 \\
            \bottomrule
        \end{tabular}
       
        }
       
      \end{sc}
    \end{small}
\end{table}

%%%%%%%%%%%%%%%%%%%%%%%%%%%%%%%%%%%%%%%%%%%%%%%%%%%%%%%%%%%%%%%%%%%%%%%%

\subsection{Expressivity Analysis}\label{sec:expressiveness}
The expressive power of graph Transformers compared to classic GNNs has hardly been studied, since the soft structural inductive bias introduced in absolute encoding is generally hard to characterize. Thanks to the unique design of our SAT, which relies on a subgraph structure extractor, it becomes possible to study the expressiveness of the output representations. More specifically, we formally show that the node representation from a structure-aware attention layer is at least as expressive as its subgraph representation given by the structure extractor, following the injectivity of the attention function with respect to the query:
\begin{theorem}\label{th:expressive}
    Assume that the space of node attributes $\Xcal$ is countable. For any pair of nodes $v$ and $v'$ in two graphs $G=(V,E,\X)$ and $G'=(V',E',\X')$, assume that there exist a node $u_1$ in $V$ such that $x_{u_1}\neq x_{w}$ for any $w\in V$ and a node $u_2$ in $V$ such that its subgraph representation $\varphi(u_2,G)\neq \varphi(w,G)$ for any $w\in V$. Then, there exists a set of parameters and a mapping $f:\Xcal\to \real^{d_{out}}$ such that their representations after the structure-aware attention are different, \ie $\text{SA-attn}(v)\neq \text{SA-attn}(v')$, if their subgraph representations are different, \ie $\varphi(v,G)\neq \varphi(v', G')$.
\end{theorem}

Note that the assumptions made in the theorem are mild as one can always add some absolute encoding or random noise to make the attributes of one node different from all other nodes, and similarly for subgraph representations. The countable assumption on $\Xcal$ is generally adopted for expressivity analysis of GNNs (\eg \citet{xu2018how}). We assume $f$ to be any mapping rather than just a linear function as in the definition of the self-attention function since it can be practically approximated by a FFN in multi-layer Transformers through the universal approximation theorem~\citep{hornik1991approximation}. Theorem~\ref{th:expressive} suggests that if the structure extractor is sufficiently expressive, the resulting SAT model can also be at least equally expressive. Furthermore, more expressive extractors could lead to more expressively powerful SAT models and thus better prediction performance, which is also empirically confirmed in Section~\ref{sec:experiments}.

%%%%%%%%%%%%%%%%%%%%%%%%%%%%%%%%%%%%%%%%%%%%%%%%%%%%%%%%%%%%%%%%%%%%%%%%
% SOTA Table
%%%%%%%%%%%%%%%%%%%%%%%%%%%%%%%%%%%%%%%%%%%%%%%%%%%%%%%%%%%%%%%%%%%%%%%%

%%%%%%%%%%%%%%%%%%%%%%%%%%%%%%%%%%%%%%%%%%%%%%%%%%%%%%%%%%%%%%%%%%%%%%%%
% Comparison to SOTA Transformers
%%%%%%%%%%%%%%%%%%%%%%%%%%%%%%%%%%%%%%%%%%%%%%%%%%%%%%%%%%%%%%%%%%%%%%%%

\begin{table}[tbp]
  \caption{Comparison of SAT to SOTA methods on OGB datasets.}
  \label{tab:comparison_to_ogb_SOTA}
  \centering
    \begin{small}
    \setlength{\tabcolsep}{0.50em}
      \begin{sc}
      
        \resizebox{.48\textwidth}{!}{%
      
            \begin{tabular}{lccc}
            \toprule
            {}&\textbf{OGBG-PPA} $\shortuparrow$ &  \textbf{OGBG-CODE2} \vspace{0.5em} $\shortuparrow$ \\
            \# graphs & 158,100 & 452,741 \\
            Avg. \# nodes & 243.4 & 125.2 \\
            Avg. \# edges & 2,266.1 & 124.2	 \\
            Metric & Accuracy & F1 score \\
            \midrule
            GCN & 0.6839$\pm$0.0084 & 0.1507$\pm$0.0018 \\
            GCN-Virtual Node & 0.6857$\pm$0.0061 & 0.1595$\pm$0.0018	\\
            GIN & 0.6892$\pm$0.0100 & 0.1495$\pm$0.0023\\
            GIN-Virtual Node & 0.7037$\pm$0.0107 & 0.1581$\pm$0.0026 \\
            \leo{DeeperGCN} & 0.7712$\pm$0.0071 & -- \\ 
            \leo{ExpC} & \textbf{0.7976$\pm$0.0072} & -- \\
            \midrule
            Transformer & 0.6454$\pm$0.0033 & 0.1670$\pm$0.0015 \\
            GraphTrans & -- & 0.1830$\pm$0.0024 \\
            \midrule
            k-subtree SAT & 0.7522$\pm$0.0056 & \textbf{0.1937$\pm$0.0028} \\
            % k-subgraph SAT & \\
            \bottomrule
        \end{tabular}
        
        }
        
      \end{sc}
    \end{small}
\end{table}
%%%%%%%%%%%%%%%%%%%%%%%%%%%%%%%%%%%%%%%%%%%%%%%%%%%%%%%%%%%%%%%%%%%%%%%%

\section{Experiments}\label{sec:experiments}
In this section, we evaluate SAT models versus several SOTA methods for graph representation learning, including GNNs and Transformers, on five graph and node prediction tasks, as well as analyze the different components of our architecture to identify what drives the performance. In summary, we discovered the following aspects about SAT:
%
%%%%%%%%%%%%%%%%%%%%%%%%%%%%%%%%%%%%%%%%%%%%%%%%%%%%%%%%%%%%%%%%%%%%%%%%
% Comparison to GNNs Table
%%%%%%%%%%%%%%%%%%%%%%%%%%%%%%%%%%%%%%%%%%%%%%%%%%%%%%%%%%%%%%%%%%%%%%%%

%%%%%%%%%%%%%%%%%%%%%%%%%%%%%%%%%%%%%%%%%%%%%%%%%%%%%%%%%%%%%%%%%%%%%%%%
% Comparison to GNN Sparse
%%%%%%%%%%%%%%%%%%%%%%%%%%%%%%%%%%%%%%%%%%%%%%%%%%%%%%%%%%%%%%%%%%%%%%%%

\begin{table*}[t]
  \caption{Since SAT uses a GNN to extract structures, we compare the performance of the original sparse GNN to SAT which uses that GNN (``base GNN"). Across different choices of GNNs, we observe that both $k$-subtree and $k$-subgraph SATs always outperform the original sparse GNN it uses. The evaluation metrics are the same as in Table~\ref{tab:comparison_to_SOTA}.}
  \label{tab:comparison_to_sparse_GNNs}
  \centering
    % \begin{small}
    \setlength{\tabcolsep}{0.50em}
      \begin{sc}
          %\comment{
          
          \resizebox{0.65\textwidth}{!}{
          
          \begin{tabular}{clcccc}
            \toprule
            {}&{}&\multicolumn{2}{c}{\textbf{ZINC}$\shortdownarrow$}&\multicolumn{1}{c}{\textbf{CLUSTER}$\shortuparrow$} &\multicolumn{1}{c}{\textbf{PATTERN}$\shortuparrow$} \\
            & & w/ edge attr. & w/o edge attr. & All & All \\
            \toprule
            {\multirow{3}{*}{\rotatebox[origin=c]{90}{GCN}}} & Base GNN & 0.192$\pm$0.015 & 0.367$\pm$0.011 & 68.498$\pm$0.976 & 71.892$\pm$0.334 \\
            & K-subtree SAT & 0.127$\pm$0.010 & \textbf{0.174$\pm$0.009} & 77.247$\pm$0.094 & 86.749$\pm$0.065 \\
            & K-subgraph SAT & \textbf{0.114$\pm$0.005} & 0.184$\pm$0.002 & \textbf{77.682$\pm$0.098} & \textbf{86.816$\pm$0.028} \\
            \bottomrule
            {\multirow{3}{*}{\rotatebox[origin=c]{90}{GIN}}} & Base GNN & 0.209$\pm$0.009 & 0.387$\pm$0.015 & 64.716$\pm$1.553 & 85.590$\pm$0.011 \\
            & K-subtree SAT & 0.115$\pm$0.005 & 0.166$\pm$0.007 & 77.255$\pm$0.085 & \textbf{86.759$\pm$0.022} \\
            & K-subgraph SAT & \textbf{0.095$\pm$0.002}& \textbf{0.162$\pm$0.013} & \textbf{77.502$\pm$0.282} & 86.746$\pm$0.014 \\
            \bottomrule
            {\multirow{3}{*}{\rotatebox[origin=c]{90}{\tiny GraphSAGE}}} & Base GNN & -- & 0.398$\pm$0.002 & 63.844$\pm$0.110 & 50.516$\pm$0.001 \\
            & K-subtree SAT & -- & \textbf{0.164$\pm$0.004} & 77.592$\pm$0.074 & 86.818$\pm$0.043 \\
            & K-subgraph SAT & -- & 0.168$\pm$0.005 & \textbf{77.657$\pm$0.185} & \textbf{86.838$\pm$0.010} \\
            \bottomrule
            {\multirow{3}{*}{\rotatebox[origin=c]{90}{PNA}}} & Base GNN & 0.188$\pm$0.004 & 0.320$\pm$0.032 & 67.077$\pm$0.977 & 86.567$\pm$0.075 \\
            & K-subtree SAT & 0.102$\pm$0.005 & 0.147$\pm$0.001 & 77.751$\pm$0.121 & \textbf{86.865$\pm$0.043} \\
            & K-subgraph SAT & \textbf{0.094$\pm$0.008} & \textbf{0.131$\pm$0.002} & \textbf{77.856$\pm$0.104} & 86.848$\pm$0.037 \\
            \bottomrule
        \end{tabular}
        
        }
        
        %}
        %}
      \end{sc}
    % \end{small}
\end{table*}

%%%%%%%%%%%%%%%%%%%%%%%%%%%%%%%%%%%%%%%%%%%%%%%%%%%%%%%%%%%%%%%%%%%%%%%%
%
\begin{compactitem}
    \item The structure-aware framework achieves SOTA performance on graph and node classification tasks, outperforming SOTA graph Transformers and sparse GNNs.
    \item Both instances of the SAT, namely $k$-subtree and $k$-subgraph SAT, \emph{always} improve upon the base GNN it is built upon, highlighting the improved expressiveness of our structure-aware approach.
    \item We show that incorporating the structure via our structure-aware attention brings a notable improvement relative to the vanilla Transformer with RWPE that just uses node attribute similarity instead of also incorporating structural similarity. We also show that a small value of $k$ already leads to good performance, while not suffering from over-smoothing or over-squashing.
    \item We show that choosing a proper absolute positional encoding and a readout method improves performance, but to a much lesser extent than incorporating the structure into the approach.
\end{compactitem}

\leo{Furthermore, we note that SAT achieves SOTA performance while only considering a small  hyperparameter search space. Performance could likely be further improved with more hyperparameter tuning.}

\subsection{Datasets and Experimental Setup}\label{sec: Implementation details}

We assess the performance of our method with five medium to large benchmark datasets for node and graph property prediction, including ZINC~\citep{dwivedi2020benchmarkgnns}, CLUSTER~\citep{dwivedi2020benchmarkgnns}, PATTERN~\citep{dwivedi2020benchmarkgnns}, OGBG-PPA~\citep{hu2020open} and OGBG-CODE2~\citep{hu2020open}. 

We compare our method to the following GNNs: GCN~\citep{Kipf2017}, GraphSAGE~\citep{hamilton2017inductive}, GAT~\citep{velickovic2018graph}, GIN~\citep{xu2018how}, PNA~\citep{corso2020principal}, \leo{DeeperGCN~ \citep{li2020deepergcn}, and ExpC~\citep{Yang2022}}. Our comparison partners also include several recently proposed Transformers on graphs, including the original Transformer with RWPE~\citep{dwivedi2022graph}, Graph Transformer~\citep{dwivedi2021generalization}, SAN~\citep{kreuzer2021rethinking}, Graphormer~\citep{ying2021do} and GraphTrans~\citep{jain2021representing}, a model that uses the vanilla Transformer on top of a GNN. 

All results for the comparison methods are either taken from the original paper or from \citet{dwivedi2020benchmarkgnns} if not available. We consider $k$-subtree and $k$-subgraph SAT equipped with different GNN extractors, including GCN, GIN, GraphSAGE and PNA. For OGBG-PPA and OGBG-CODE2, we do not run experiments for $k$-subgraph SAT models due to large memory requirements. Full details on the datasets, \leo{experimental setup, and hyperparameters} are provided in the Appendix. 

\subsection{Comparison to State-of-the-Art Methods}
We show the performance of SATs compared to other GNNs and Transformers in Table~\ref{tab:comparison_to_SOTA} and \ref{tab:comparison_to_ogb_SOTA}. SAT models consistently outperform SOTA methods on these datasets, showing its ability to combine the benefits of both GNNs and Transformers. In particular, for \leo{the CODE2} dataset, our SAT models outperform SOTA methods by a large margin despite a relatively small number of parameters and minimal hyperparameter tuning, which will put it at the first place on the \leo{OGB leaderboard}.

\subsection{SAT Models vs. Sparse GNNs}
Table~\ref{tab:comparison_to_sparse_GNNs} summarizes the performance of SAT relative to the sparse GNN it uses to extract the subgraph representations, across different GNNs. We observe that both variations of SAT consistently bring large performance gains to its base GNN counterpart, making it a systematic enhancer of any GNN model. Furthermore, PNA, which is the most  expressive GNN we considered, has consistently the best performance when used with SAT, empirically validating our theoretical finding in Section~\ref{sec:expressiveness}. $k$-subgraph SAT also outperforms or performs equally as $k$-subtree SAT in almost all the cases, showing its superior expressiveness.

\subsection{Hyperparameter Studies}
While Table~\ref{tab:comparison_to_sparse_GNNs} showcases the added value of the SAT relative to sparse GNNs, we now  dissect the components of SAT on the ZINC dataset to identify which aspects of the architecture bring the biggest performance gains. 

\paragraph{Effect of $k$ in SAT}

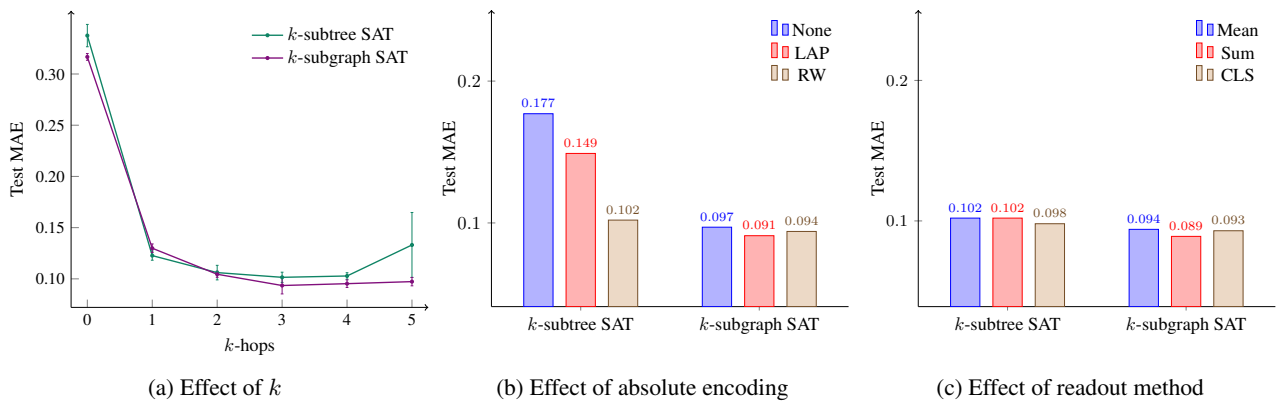
\begin{figure*}[t]
    \pgfplotsset{
        every non boxed y axis/.append style={y axis line style=->}
        }
     \centering
     \resizebox{1.\textwidth}{!}{
     \begin{subfigure}[b]{0.33\textwidth}
         \centering
         %\begin{figure}[h!]
    % add arrows to axes
    \pgfplotsset{
        every x axis/.append style={x axis line style=->},
        every non boxed y axis/.append style={y axis line style=->}
        }
    \centering
    \begin{tikzpicture}[scale=0.7]
      \pgfplotsset{
        every axis/.append style = {
          mark size = 0.75pt, 
        }
      }
      \begin{axis}[%
        axis x line*      = bottom,
        axis y line*      = left,
        axis line style   = ->,
        %enlargelimits     = true,
        %xmin              = 225.0,
        %xmax              = 1000.0,
        %ymin              = 0,
        %ymax              = 85,
        xlabel            = {$k$-hops},
        ylabel            = {Test MAE},
        ticklabel style={font=\small},
        xtick             = {0,1,2,3,4,5},
        xticklabels       = {0,1,2,3,4,5},
        ytick             = {0.10,0.15,0.20,0.25,0.30},
        yticklabels       = {0.10,0.15,0.20,0.25,0.30},
        label style       = {font={\normalsize}},
        legend pos        = {north east},
        legend cell align = left,
        legend style      = {%
          draw = none,
        enlarge x limits  = 0.05,
        enlarge y limits  = 0.05
        },
        legend image post style = {
          scale = 1.0
        },
        legend columns=1, 
        legend style={
            font={\normalsize},
                    % the /tikz/ prefix is necessary here...
                    % otherwise, it might end-up with `/pgfplots/column 2`
                    % which is not what we want. compare pgfmanual.pdf
            /tikz/column 2/.style={
                column sep=5pt,
            },
        },
      ]
         % adds a line or a points
          \addplot[emerald, mark=*, line width = 0.75pt, error bars/.cd, y fixed, y dir=both, y explicit] table[col sep = comma, x index = 0, y index = 3, y error index = 4]{data/khop.csv};
          \addplot[538red, mark=*, line width = 0.75pt, error bars/.cd, y fixed, y dir=both, y explicit] table[col sep = comma, x index = 0, y index = 1, y error index = 2] {data/khop.csv};
          
          \legend{$k$-subtree SAT, $k$-subgraph SAT};
      \end{axis}
  \end{tikzpicture}
    \caption{Effect of $k$}
    %\caption{Test MAE for different values of $k$ on the $k$-subtree and $k$-subgraph SAT (lower is better) on the ZINC dataset.}
    \label{fig:effect_of_k}
%\end{figure}
     \end{subfigure}
     \hfill
     \begin{subfigure}[b]{0.33\textwidth}
         \centering
         %\begin{figure}
\centering

\begin{tikzpicture}[scale=0.7]
 
\begin{axis} [ybar = .25cm,
    yshift            = 5mm,  
    ylabel shift = -2 mm,
    bar width = 16pt,
    axis lines = left,
    xmin = 0.5, 
    xmax = 2.5,
    ymax = 0.2,
    label style       = {font={\normalsize}},
    ticklabel style={font=\normalsize},
    ylabel={Test MAE},
    xlabel={\textcolor{white}{x}},
    enlarge y limits = {abs = .05},
    ytick = {0.10,0.20},
    xtick = {1,2},
    xticklabels={\normalsize $k$-subtree SAT, \normalsize $k$-subgraph SAT},
    nodes near coords,
    every node near coord/.append style={font=\scriptsize},
    nodes near coords style={/pgf/number format/.cd,fixed, zerofill,precision=3},
    legend style={font=\normalsize, draw=none},
]
\addplot coordinates {(1,0.177) (2,0.097)}; % None
\addplot coordinates {(1,0.149) (2,0.091)}; % LAP
\addplot coordinates {(1,0.102) (2,0.094)}; % RW
\legend{None, LAP, RW}
\end{axis}

\end{tikzpicture}
\caption{Effect of absolute encoding}
\label{fig:pos_enc}
%\end{figure}
     \end{subfigure}
     \hfill
     \begin{subfigure}[b]{0.33\textwidth}
         \centering
         %\begin{figure}
\centering

\begin{tikzpicture}[scale=0.7]
 
\begin{axis} [ybar = .25cm,
    bar width = 16pt,
    ylabel shift = -2 mm,
    axis lines = left,
    xmin = 0.5, 
    xmax = 2.5,
    ymax = 0.2,
    label style       = {font={\normalsize}},
    ticklabel style={font=\small},
    ylabel={Test MAE},
    xlabel={\textcolor{white}{x}},
    enlarge y limits = {abs = .05},
    ytick = {0.10,0.20},
    xtick = {1,2},
    xticklabels={\normalsize $k$-subtree SAT, \normalsize $k$-subgraph SAT},
    nodes near coords,
    every node near coord/.append style={font=\scriptsize},
    nodes near coords style={/pgf/number format/.cd,fixed, zerofill,precision=3},
    legend style={font=\normalsize, draw=none},
]
\addplot coordinates {(1,0.102) (2,0.094)}; % None
\addplot coordinates {(1,0.102) (2,0.089)}; % LAP
\addplot coordinates {(1,0.098) (2,0.093)}; % RW
\legend{Mean, Sum, CLS}
\end{axis}

\end{tikzpicture}
\caption{Effect of readout method}
\label{fig:readout}
%\end{figure}
     \end{subfigure}
     }
        \caption{We provide an analysis of the different drivers of performance in SAT on the ZINC dataset (lower is better). In Figure~\ref{fig:effect_of_k}, we show how changing the size of $k$ affects performance ($k$=0 is equivalent to a vanilla Transformer that is not structure-aware). Figure~\ref{fig:pos_enc} shows the effect of different absolute encoding methods, and Figure~\ref{fig:readout} shows the effect of different readout methods.}
        \label{fig:ablation}
\end{figure*}

The key contribution of SAT is its ability to explicitly incorporate structural information in the self-attention. Here, we seek to demonstrate that this information provides crucial predictive information, and study how the choice of $k$ affects the results. Figure~\ref{fig:effect_of_k} shows how the test MAE is impacted by varying $k$ for $k$-subtree and $k$-subgraph extractors using PNA on the ZINC dataset. All models use the RWPE. $k=0$ corresponds to the vanilla Transformer only using absolute positional encoding, \ie not using structure. We find that incorporating structural information leads to substantial improvement in performance, with optimal performance around $k=3$ for both $k$-subtree and $k$-subgraph extractors. As $k$ increases beyond $k=4$, the performance in $k$-subtree extractors deteriorated, which is consistent with the observed phenomenon that GNNs work best in shallower networks~\citep{Kipf2017}. We observe that $k$-subgraph does not suffer as much from this issue, underscoring a new aspect of its usefulness. On the other hand, $k$-subtree extractors are more computationally efficient and scalable to larger OGB datasets.

\paragraph{Effect of absolute encoding}

%%%%%%%%%%%%%%%%%%%%%%%%%%%%%%%%%%%%%%%%%%%%%%%%%%%%%%%%%%%%%
%%%%%%%%%%%%%%%%%%%%%%%%%%%%%%%%%%%%%%%%%%%%%%%%%%%%%%%%%%%%%

We assess here whether the absolute encoding brought complementary information to SAT. In Figure~\ref{fig:pos_enc}, we conduct an ablation study showing the results of SAT with and without absolute positional encoding, including RWPE and Laplacian PE~\citep{dwivedi2020benchmarkgnns}. Our SAT with a positional encoding outperforms its counterpart without it, confirming the complementary nature of the two encodings. However, we also note that the performance gain brought by the absolute encoding is far less than the gain obtained by using our structure-aware attention, as shown in  Figure~\ref{fig:effect_of_k} (\leo{comparing the instance of $k=0$ to $k>0$}), emphasizing that our structure-aware attention is the more important aspect of the model.

\paragraph{Comparison of readout methods}

Finally, we compare the performance of SAT models using different readout methods for aggregating node-level representations on the ZINC dataset in Figure~\ref{fig:readout}, including the CLS pooling discussed in Section~\ref{sec:pooling}. Unlike the remarkable influence of the readout method in GNNs~\citep{xu2018how}, we observe very little impact in SAT models.

\subsection{Model Interpretation}
\begin{figure}[h!]
    \includegraphics[width=.3\columnwidth]{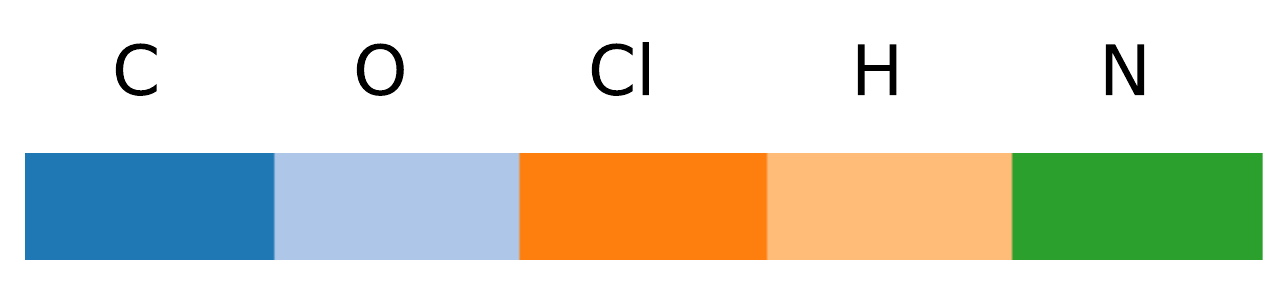} \\
    \includegraphics[width=.98\columnwidth]{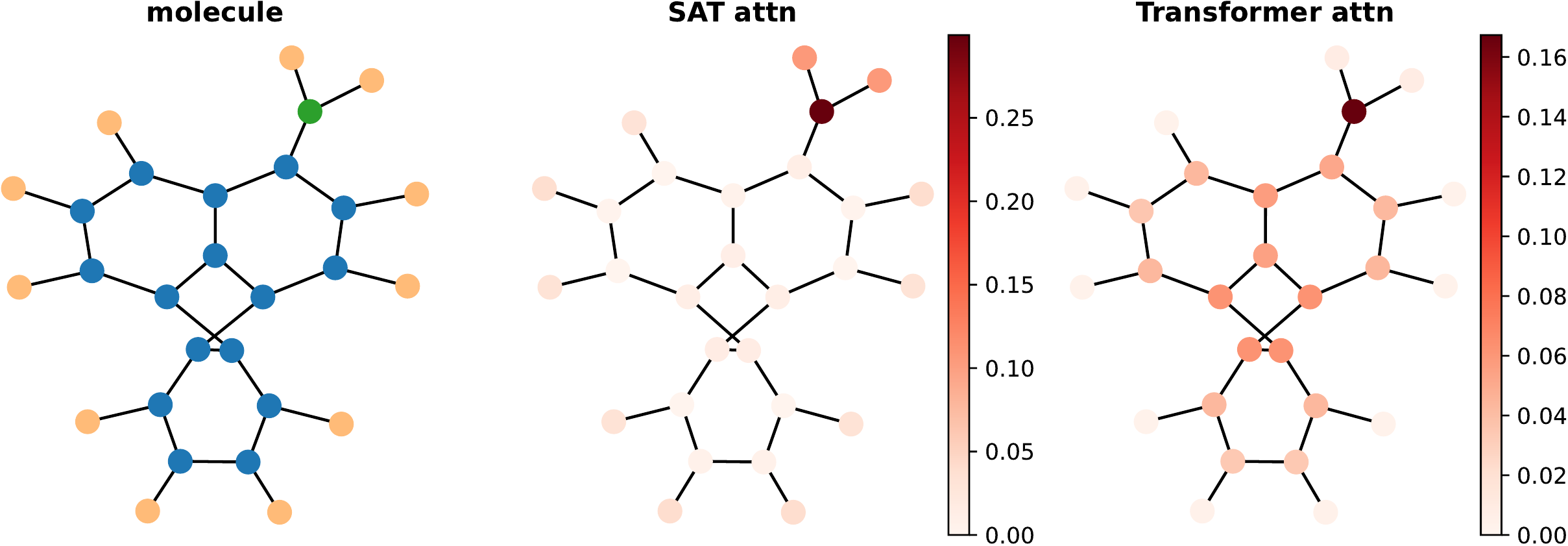}
    \caption{Attention visualization of SAT and the Transformer. The center column shows the attention weights of the [CLS] node learned by our SAT model and the right column shows the attention weights learned by the classic Transformer with the random walk positional encoding (RWPE).}
    \label{fig:attn_score}
\end{figure}

In addition to performance improvement, we show that SAT offers better model interpretability compared to the classic Transformer with only absolute positional encoding. We respectively train a SAT model and a Transformer with a CLS readout on the Mutagenicity dataset, and visualize the attention scores between the [CLS] node and other nodes learned by SAT and the Transformer in Figure~\ref{fig:attn_score}. \leo{The salient difference between the two models is that SAT has structure-aware node embeddings, and thus we can attribute the following interpretability gains to that.} While both models manage to identify some chemical motifs known for mutagenicity, such as NO$_2$ and NH$_2$, the attention scores learned by SAT are sparser and more informative, \leo{meaning that SAT puts more attention weights on these known mutagenic motifs than the Transformer with RWPE.} The vanilla Transformer even fails to put attention on some important atoms such as the H atoms in the NH$_2$ group. The only H atoms highlighted by SAT are those in the NH$_2$ group, suggesting that our SAT indeed takes the structure into account.   \leo{More focus on these discriminative motifs makes the SAT model less influenced by other chemical patterns that commonly exist in the dataset, such as benzene, and thus leads to overall improved performance.} More results are provided in the Appendix.

\section{Discussion}
We introduced the SAT model, which successfully incorporates structural information into the Transformer architecture and overcomes the limitations of the absolute encoding. In addition to SOTA empirical performance  \leo{with minimal hyperparameter tuning}, SAT also provides better interpretability than the Transformer. 

\paragraph{Limitations}
As mentioned above, $k$-subgraph SAT has higher memory requirements than $k$-subtree SAT, which can restrict its applicability if access to high memory GPUs is restricted. We see the main limitation of SAT is that it suffers from the same drawbacks as the Transformer, namely the quadratic complexity of the self-attention computation.%, but will benefit from the recent line of research on efficient Transformers~\citep{tay2020efficient}. 

\leo{
\paragraph{Future work} Because SAT can be combined with any GNN, a natural extension of our work is to combine SAT with structure extractors which have shown to be strictly more expressive than the 1-WL test, such as the recent topological GNN introduced by \citet{Horn21a}. Additionally, the SAT framework is flexible and can incorporate any structure extractor which produces structure-aware node representations, and could even be extended beyond using GNNs, such as differentiable graph kernels. 

Another important area for future work is to focus on reducing the high memory cost and time complexity of the self-attention computation, as is being done in recent efforts for developing a so-called linear transformer, which has linear complexity in both time and space requirements \citep{tay2020efficient, linformer, zhen2022cosformer}.
}

\section*{Acknowledgements}
This work was supported in part by the Alfried Krupp Prize for Young University Teachers of the Alfried Krupp von Bohlen und Halbach-Stiftung (K.B.). The authors would also like to thank Dr. Bastian Rieck and Dr. Carlos Oliver for their insightful feedback on the manuscript, which greatly improved it.

\bibliography{mybib}

\begin{thebibliography}{55}
\providecommand{\natexlab}[1]{#1}
\providecommand{\url}[1]{\texttt{#1}}
\expandafter\ifx\csname urlstyle\endcsname\relax
  \providecommand{\doi}[1]{doi: #1}\else
  \providecommand{\doi}{doi: \begingroup \urlstyle{rm}\Url}\fi

\bibitem[Abbe(2018)]{PatternCluster2018}
Abbe, E.
\newblock Community detection and stochastic block models: Recent developments.
\newblock \emph{Journal of Machine Learning Research (JMLR)}, 18\penalty0
  (177):\penalty0 1--86, 2018.

\bibitem[Alon \& Yahav(2021)Alon and Yahav]{alon2021on}
Alon, U. and Yahav, E.
\newblock On the bottleneck of graph neural networks and its practical
  implications.
\newblock In \emph{International Conference on Learning Representations
  (ICLR)}, 2021.

\bibitem[Alsentzer et~al.(2020)Alsentzer, Finlayson, Li, and
  Zitnik]{alsentzer2020subgraph}
Alsentzer, E., Finlayson, S.~G., Li, M.~M., and Zitnik, M.
\newblock Subgraph neural networks.
\newblock In \emph{Proceedings of Neural Information Processing Systems,
  NeurIPS}, 2020.

\bibitem[Bodnar et~al.(2021)Bodnar, Frasca, Otter, Wang, Li{\`o}, Montufar, and
  Bronstein]{bodnar2021weisfeiler}
Bodnar, C., Frasca, F., Otter, N., Wang, Y.~G., Li{\`o}, P., Montufar, G.~F.,
  and Bronstein, M.
\newblock Weisfeiler and lehman go cellular: Cw networks.
\newblock In \emph{Advances in Neural Information Processing Systems
  (NeurIPS)}, 2021.

\bibitem[Chen et~al.(2020)Chen, Lin, Li, Li, Zhou, and Sun]{chen2020measuring}
Chen, D., Lin, Y., Li, W., Li, P., Zhou, J., and Sun, X.
\newblock Measuring and relieving the over-smoothing problem for graph neural
  networks from the topological view.
\newblock In \emph{Proceedings of the AAAI Conference on Artificial
  Intelligence}, 2020.

\bibitem[Corso et~al.(2020)Corso, Cavalleri, Beaini, Li{\`o}, and
  Veli{\v{c}}kovi{\'c}]{corso2020principal}
Corso, G., Cavalleri, L., Beaini, D., Li{\`o}, P., and Veli{\v{c}}kovi{\'c}, P.
\newblock Principal neighbourhood aggregation for graph nets.
\newblock In \emph{Advances in Neural Information Processing Systems
  (NeurIPS)}, 2020.

\bibitem[Dosovitskiy et~al.(2020)Dosovitskiy, Beyer, Kolesnikov, Weissenborn,
  Zhai, Unterthiner, Dehghani, Minderer, Heigold, Gelly,
  et~al.]{dosovitskiy2020image}
Dosovitskiy, A., Beyer, L., Kolesnikov, A., Weissenborn, D., Zhai, X.,
  Unterthiner, T., Dehghani, M., Minderer, M., Heigold, G., Gelly, S., et~al.
\newblock An image is worth 16x16 words: Transformers for image recognition at
  scale.
\newblock In \emph{International Conference on Learning Representations
  (ICLR)}, 2020.

\bibitem[Dwivedi \& Bresson(2021)Dwivedi and
  Bresson]{dwivedi2021generalization}
Dwivedi, V.~P. and Bresson, X.
\newblock A generalization of transformer networks to graphs.
\newblock In \emph{AAAI Workshop on Deep Learning on Graphs: Methods and
  Applications}, 2021.

\bibitem[Dwivedi et~al.(2020)Dwivedi, Joshi, Laurent, Bengio, and
  Bresson]{dwivedi2020benchmarkgnns}
Dwivedi, V.~P., Joshi, C.~K., Laurent, T., Bengio, Y., and Bresson, X.
\newblock Benchmarking graph neural networks.
\newblock \emph{arXiv preprint arXiv:2003.00982}, 2020.

\bibitem[Dwivedi et~al.(2022)Dwivedi, Luu, Laurent, Bengio, and
  Bresson]{dwivedi2022graph}
Dwivedi, V.~P., Luu, A.~T., Laurent, T., Bengio, Y., and Bresson, X.
\newblock Graph neural networks with learnable structural and positional
  representations.
\newblock In \emph{International Conference on Learning Representations}, 2022.

\bibitem[Fan et~al.(2019)Fan, Ma, Li, He, Zhao, Tang, and Yin]{fan2019graph}
Fan, W., Ma, Y., Li, Q., He, Y., Zhao, E., Tang, J., and Yin, D.
\newblock Graph neural networks for social recommendation.
\newblock In \emph{The World Wide Web Conference}, 2019.

\bibitem[Gao \& Pavel(2017)Gao and Pavel]{gao2017properties}
Gao, B. and Pavel, L.
\newblock On the properties of the softmax function with application in game
  theory and reinforcement learning.
\newblock \emph{arXiv preprint arXiv:1704.00805}, 2017.

\bibitem[Gao \& Ji(2019)Gao and Ji]{gao2019graph}
Gao, H. and Ji, S.
\newblock Graph u-nets.
\newblock In \emph{International Conference on Machine Learning}, pp.\
  2083--2092, 2019.

\bibitem[Gaudelet et~al.(2021)Gaudelet, Day, Jamasb, Soman, Regep, Liu, Hayter,
  Vickers, Roberts, Tang, et~al.]{gaudelet2021utilizing}
Gaudelet, T., Day, B., Jamasb, A.~R., Soman, J., Regep, C., Liu, G., Hayter,
  J.~B., Vickers, R., Roberts, C., Tang, J., et~al.
\newblock Utilizing graph machine learning within drug discovery and
  development.
\newblock \emph{Briefings in Bioinformatics}, 22\penalty0 (6):\penalty0
  bbab159, 2021.

\bibitem[Gilmer et~al.(2017)Gilmer, Schoenholz, Riley, Vinyals, and
  Dahl]{gilmer2017neural}
Gilmer, J., Schoenholz, S.~S., Riley, P.~F., Vinyals, O., and Dahl, G.~E.
\newblock Neural message passing for quantum chemistry.
\newblock In \emph{International Conference on Machine Learning (ICML)}, 2017.

\bibitem[Hamilton et~al.(2017)Hamilton, Ying, and
  Leskovec]{hamilton2017inductive}
Hamilton, W.~L., Ying, R., and Leskovec, J.
\newblock Inductive representation learning on large graphs.
\newblock In \emph{Advances in Neural Information Processing Systems
  (NeurIPS)}, 2017.

\bibitem[Horn et~al.(2021)Horn, {De Brouwer}, Moor, Moreau, Rieck, and
  Borgwardt]{Horn21a}
Horn, M., {De Brouwer}, E., Moor, M., Moreau, Y., Rieck, B., and Borgwardt, K.
\newblock Topological graph neural networks.
\newblock 2021.

\bibitem[Hornik(1991)]{hornik1991approximation}
Hornik, K.
\newblock Approximation capabilities of multilayer feedforward networks.
\newblock \emph{Neural networks}, 4\penalty0 (2):\penalty0 251--257, 1991.

\bibitem[Hu et~al.(2020{\natexlab{a}})Hu, Fey, Zitnik, Dong, Ren, Liu, Catasta,
  and Leskovec]{hu2020open}
Hu, W., Fey, M., Zitnik, M., Dong, Y., Ren, H., Liu, B., Catasta, M., and
  Leskovec, J.
\newblock Open graph benchmark: Datasets for machine learning on graphs.
\newblock In \emph{Advances in Neural Information Processing Systems
  (NeurIPS)}, 2020{\natexlab{a}}.

\bibitem[Hu et~al.(2020{\natexlab{b}})Hu, Liu, Gomes, Zitnik, Liang, Pande, and
  Leskovec]{hu2020pretraining}
Hu, W., Liu, B., Gomes, J., Zitnik, M., Liang, P., Pande, V., and Leskovec, J.
\newblock Strategies for pre-training graph neural networks.
\newblock In \emph{International Conference on Learning Representations
  (ICLR)}, 2020{\natexlab{b}}.

\bibitem[Ingraham et~al.(2019)Ingraham, Garg, Barzilay, and
  Jaakkola]{ingraham2019generative}
Ingraham, J., Garg, V., Barzilay, R., and Jaakkola, T.
\newblock Generative models for graph-based protein design.
\newblock In \emph{Advances in Neural Information Processing Systems
  (NeurIPS)}, 2019.

\bibitem[Irwin et~al.(2012)Irwin, Sterling, Mysinger, Bolstad, and
  Coleman]{ZINC}
Irwin, J.~J., Sterling, T., Mysinger, M.~M., Bolstad, E.~S., and Coleman, R.~G.
\newblock Zinc: A free tool to discover chemistry for biology.
\newblock \emph{Journal of Chemical Information and Modeling}, 52\penalty0
  (7):\penalty0 1757--1768, 2012.

\bibitem[Jain et~al.(2021)Jain, Wu, Wright, Mirhoseini, Gonzalez, and
  Stoica]{jain2021representing}
Jain, P., Wu, Z., Wright, M., Mirhoseini, A., Gonzalez, J.~E., and Stoica, I.
\newblock Representing long-range context for graph neural networks with global
  attention.
\newblock In \emph{Advances in Neural Information Processing Systems
  (NeurIPS)}, 2021.

\bibitem[Kersting et~al.(2016)Kersting, Kriege, Morris, Mutzel, and
  Neumann]{KKMMN2016}
Kersting, K., Kriege, N.~M., Morris, C., Mutzel, P., and Neumann, M.
\newblock Benchmark data sets for graph kernels, 2016.
\newblock \url{http://graphkernels.cs.tu-dortmund.de}.

\bibitem[Kipf \& Welling(2017)Kipf and Welling]{Kipf2017}
Kipf, T.~N. and Welling, M.
\newblock Semi-supervised classification with graph convolutional networks.
\newblock In \emph{International Conference on Learning Representations
  (ICLR)}, 2017.

\bibitem[Kreuzer et~al.(2021)Kreuzer, Beaini, Hamilton, L{\'e}tourneau, and
  Tossou]{kreuzer2021rethinking}
Kreuzer, D., Beaini, D., Hamilton, W.~L., L{\'e}tourneau, V., and Tossou, P.
\newblock Rethinking graph transformers with spectral attention.
\newblock In \emph{Advances in Neural Information Processing Systems
  (NeurIPS)}, 2021.

\bibitem[Li et~al.(2019)Li, Müller, Thabet, and Ghanem]{li2019deepgcns}
Li, G., Müller, M., Thabet, A., and Ghanem, B.
\newblock Deepgcns: Can gcns go as deep as cnns?
\newblock In \emph{Proceedings of the International Conference on Computer
  Vision (ICCV)}, 2019.

\bibitem[Li et~al.(2020{\natexlab{a}})Li, Xiong, Thabet, and
  Ghanem]{li2020deepergcn}
Li, G., Xiong, C., Thabet, A., and Ghanem, B.
\newblock Deepergcn: All you need to train deeper gcns, 2020{\natexlab{a}}.

\bibitem[Li et~al.(2020{\natexlab{b}})Li, Wang, Wang, and
  Leskovec]{li2020distance}
Li, P., Wang, Y., Wang, H., and Leskovec, J.
\newblock Distance encoding: Design provably more powerful neural networks for
  graph representation learning.
\newblock In \emph{Advances in Neural Information Processing Systems
  (NeurIPS)}, 2020{\natexlab{b}}.

\bibitem[Li et~al.(2018)Li, Han, and Wu]{Li2018}
Li, Q., Han, Z., and Wu, X.
\newblock Deeper insights into graph convolutional networks for semi-supervised
  learning.
\newblock In \emph{Proceedings of the AAAI Conference on Artificial
  Intelligence}, 2018.

\bibitem[Loshchilov \& Hutter(2016)Loshchilov and Hutter]{loshchilov2016sgdr}
Loshchilov, I. and Hutter, F.
\newblock Sgdr: Stochastic gradient descent with warm restarts.
\newblock In \emph{International Conference on Learning Representations
  (ICLR)}, 2016.

\bibitem[Loshchilov \& Hutter(2018)Loshchilov and
  Hutter]{loshchilov2018decoupled}
Loshchilov, I. and Hutter, F.
\newblock Decoupled weight decay regularization.
\newblock In \emph{International Conference on Learning Representations
  (ICLR)}, 2018.

\bibitem[Mesquita et~al.(2020)Mesquita, Souza, and Kaski]{rethinkpooling2020}
Mesquita, D., Souza, A.~H., and Kaski, S.
\newblock Rethinking pooling in graph neural networks.
\newblock In \emph{Advances in Neural Information Processing Systems
  (NeurIPS)}, 2020.

\bibitem[Mialon et~al.(2021)Mialon, Chen, Selosse, and
  Mairal]{mialon2021graphit}
Mialon, G., Chen, D., Selosse, M., and Mairal, J.
\newblock Graphit: Encoding graph structure in transformers, 2021.

\bibitem[Micchelli et~al.(2006)Micchelli, Xu, and
  Zhang]{micchelli2006universal}
Micchelli, C.~A., Xu, Y., and Zhang, H.
\newblock Universal kernels.
\newblock \emph{Journal of Machine Learning Research (JMLR)}, 7\penalty0 (12),
  2006.

\bibitem[Morris et~al.(2019)Morris, Ritzert, Fey, Hamilton, Lenssen, Rattan,
  and Grohe]{morris2019weisfeiler}
Morris, C., Ritzert, M., Fey, M., Hamilton, W.~L., Lenssen, J.~E., Rattan, G.,
  and Grohe, M.
\newblock Weisfeiler and leman go neural: Higher-order graph neural networks.
\newblock In \emph{Proceedings of the AAAI Conference on Artificial
  Intelligence}, 2019.

\bibitem[Nikolentzos \& Vazirgiannis(2020)Nikolentzos and
  Vazirgiannis]{nikolentzos2020random}
Nikolentzos, G. and Vazirgiannis, M.
\newblock Random walk graph neural networks.
\newblock In \emph{Advances in Neural Information Processing Systems
  (NeurIPS)}, 2020.

\bibitem[Oono \& Suzuki(2020)Oono and Suzuki]{oono2020graph}
Oono, K. and Suzuki, T.
\newblock Graph neural networks exponentially lose expressive power for node
  classification.
\newblock In \emph{International Conference on Learning Representations
  (ICLR)}, 2020.

\bibitem[Qin et~al.(2022)Qin, Sun, Deng, Li, Wei, Lv, Yan, Kong, and
  Zhong]{zhen2022cosformer}
Qin, Z., Sun, W., Deng, H., Li, D., Wei, Y., Lv, B., Yan, J., Kong, L., and
  Zhong, Y.
\newblock cosformer: Rethinking softmax in attention.
\newblock In \emph{International Conference on Learning Representations}, 2022.

\bibitem[Rives et~al.(2021)Rives, Meier, Sercu, Goyal, Lin, Liu, Guo, Ott,
  Zitnick, Ma, et~al.]{rives2021biological}
Rives, A., Meier, J., Sercu, T., Goyal, S., Lin, Z., Liu, J., Guo, D., Ott, M.,
  Zitnick, C.~L., Ma, J., et~al.
\newblock Biological structure and function emerge from scaling unsupervised
  learning to 250 million protein sequences.
\newblock \emph{Proceedings of the National Academy of Sciences}, 118\penalty0
  (15), 2021.

\bibitem[Rong et~al.(2020)Rong, Bian, Xu, Xie, Wei, Huang, and Huang]{Rong2020}
Rong, Y., Bian, Y., Xu, T., Xie, W., Wei, Y., Huang, W., and Huang, J.
\newblock Self-supervised graph transformer on large-scale molecular data.
\newblock In \emph{Advances in Neural Information Processing Systems
  (NeurIPS)}, 2020.

\bibitem[Shaw et~al.(2018)Shaw, Uszkoreit, and Vaswani]{shaw2018self}
Shaw, P., Uszkoreit, J., and Vaswani, A.
\newblock Self-attention with relative position representations.
\newblock In \emph{Proceedings of the North American Chapter of the Association
  for Computational Linguistics (NAACL)}, 2018.

\bibitem[Shi et~al.(2021)Shi, Huang, Feng, Zhong, Wang, and Sun]{Shi2021}
Shi, Y., Huang, Z., Feng, S., Zhong, H., Wang, W., and Sun, Y.
\newblock Masked label prediction: Unified message passing model for
  semi-supervised classification.
\newblock In Zhou, Z.-H. (ed.), \emph{Proceedings of the Thirtieth
  International Joint Conference on Artificial Intelligence~(IJCAI-21)}, pp.\
  1548--1554. International Joint Conferences on Artificial Intelligence
  Organization, 8 2021.

\bibitem[Tay et~al.(2020)Tay, Dehghani, Bahri, and Metzler]{tay2020efficient}
Tay, Y., Dehghani, M., Bahri, D., and Metzler, D.
\newblock Efficient transformers: A survey.
\newblock \emph{arXiv preprint arXiv:2009.06732}, 2020.

\bibitem[Vaswani et~al.(2017)Vaswani, Shazeer, Parmar, Uszkoreit, Jones, Gomez,
  Kaiser, and Polosukhin]{vaswani2017attention}
Vaswani, A., Shazeer, N., Parmar, N., Uszkoreit, J., Jones, L., Gomez, A.~N.,
  Kaiser, {\L}., and Polosukhin, I.
\newblock Attention is all you need.
\newblock In \emph{Advances in Neural Information Processing Systems
  (NeurIPS)}, 2017.

\bibitem[Veli{\v{c}}kovi{\'{c}} et~al.(2018)Veli{\v{c}}kovi{\'{c}}, Cucurull,
  Casanova, Romero, Li{\`{o}}, and Bengio]{velickovic2018graph}
Veli{\v{c}}kovi{\'{c}}, P., Cucurull, G., Casanova, A., Romero, A., Li{\`{o}},
  P., and Bengio, Y.
\newblock {Graph Attention Networks}.
\newblock In \emph{International Conference on Learning Representations
  (ICLR)}, 2018.

\bibitem[Wang et~al.(2020)Wang, Li, Khabsa, Fang, and Ma]{linformer}
Wang, S., Li, B.~Z., Khabsa, M., Fang, H., and Ma, H.
\newblock Linformer: Self-attention with linear complexity, 2020.

\bibitem[Wijesinghe \& Wang(2022)Wijesinghe and Wang]{wijesinghe2022a}
Wijesinghe, A. and Wang, Q.
\newblock A new perspective on ''how graph neural networks go beyond
  weisfeiler-lehman?''.
\newblock In \emph{International Conference on Learning Representations}, 2022.

\bibitem[Xu et~al.(2019)Xu, Hu, Leskovec, and Jegelka]{xu2018how}
Xu, K., Hu, W., Leskovec, J., and Jegelka, S.
\newblock How powerful are graph neural networks?
\newblock In \emph{International Conference on Learning Representations
  (ICLR)}, 2019.

\bibitem[Yang et~al.(2022)Yang, Wang, Shen, Qi, and Yin]{Yang2022}
Yang, M., Wang, R., Shen, Y., Qi, H., and Yin, B.
\newblock Breaking the expression bottleneck of graph neural networks.
\newblock \emph{IEEE Transactions on Knowledge and Data Engineering}, pp.\
  1--1, 2022.
\newblock \doi{10.1109/TKDE.2022.3168070}.

\bibitem[Ying et~al.(2021)Ying, Cai, Luo, Zheng, Ke, He, Shen, and
  Liu]{ying2021do}
Ying, C., Cai, T., Luo, S., Zheng, S., Ke, G., He, D., Shen, Y., and Liu, T.-Y.
\newblock Do transformers really perform badly for graph representation?
\newblock In \emph{Advances in Neural Information Processing Systems
  (NeurIPS)}, 2021.

\bibitem[Ying et~al.(2018)Ying, You, Morris, Ren, Hamilton, and
  Leskovec]{ying2018hierarchical}
Ying, Z., You, J., Morris, C., Ren, X., Hamilton, W., and Leskovec, J.
\newblock Hierarchical graph representation learning with differentiable
  pooling.
\newblock \emph{Advances in neural information processing systems}, 31, 2018.

\bibitem[You et~al.(2019)You, Ying, and Leskovec]{you2019position}
You, J., Ying, R., and Leskovec, J.
\newblock Position-aware graph neural networks.
\newblock In \emph{International Conference on Machine Learning (ICML)}, 2019.

\bibitem[Zhang et~al.(2020)Zhang, Zhang, Xia, and Sun]{zhang2020graph}
Zhang, J., Zhang, H., Xia, C., and Sun, L.
\newblock Graph-bert: Only attention is needed for learning graph
  representations.
\newblock \emph{arXiv preprint arXiv:2001.05140}, 2020.

\bibitem[Zhang \& Li(2021)Zhang and Li]{zhang2021nested}
Zhang, M. and Li, P.
\newblock Nested graph neural networks.
\newblock In \emph{Proceedings of the 35th Conference on Neural Information
  Processing Systems (NeurIPS)}, 2021.

\end{thebibliography}

\bibliographystyle{icml2022}

%%%%%%%%%%%%%%%%%%%%%%%%%%%%%%%%%%%%%%%%%%%%%%%%%%%%%%%%%%%%%%%%%%%%%%%%%%%%%%%
%%%%%%%%%%%%%%%%%%%%%%%%%%%%%%%%%%%%%%%%%%%%%%%%%%%%%%%%%%%%%%%%%%%%%%%%%%%%%%%
% APPENDIX
%%%%%%%%%%%%%%%%%%%%%%%%%%%%%%%%%%%%%%%%%%%%%%%%%%%%%%%%%%%%%%%%%%%%%%%%%%%%%%%
%%%%%%%%%%%%%%%%%%%%%%%%%%%%%%%%%%%%%%%%%%%%%%%%%%%%%%%%%%%%%%%%%%%%%%%%%%%%%%%
\newpage
\appendix
\onecolumn

\vspace*{0.3cm}
\begin{center}
    {\huge Appendix}
\end{center}
\vspace*{0.5cm}

\setcounter{theorem}{0}
This appendix provides both theoretical and experimental materials and is organized as follows: Section~\ref{sec:supp_background} provides a more detailed background on graph neural networks. Section~\ref{sec:supp_theory} presents proofs of Theorem 1 and 2. Section~\ref{sec:supp_experiments} provides experimental details and additional results. Section~\ref{sec:supp_interpretation} provides details on the model interpretation and additional visualization results.

\section{Background on Graph Neural Networks}\label{sec:supp_background}
The overarching idea of a graph neural network is to iteratively update a node's embedding by incorporating information sent from its neighbors. \citet{xu2018how} provide a general framework of the steps incorporated in this process by generalizing the different frameworks into $\AGGREGATE$, $\COMBINE$ and $\READOUT$ steps. The various flavors of GNNs can be typically understood as variations within these three functions. For a given layer $l$, the $\AGGREGATE$ step aggregates (e.g. using the sum or mean) the representations of the neighbors of a given node, which is then combined with the given node's representation from the previous layer in the $\COMBINE$ step. This is followed by a non-linear function, such as $\mathrm{ReLU}$, and the updated node representations are then passed to the next layer. These two steps are repeated for as many layers as there are in the network. It is worth noting that the output of these two steps provides representations of nodes which accounts for local sub-structures of size only increased by one, which would thus require a very deep network to capture interactions between the given node and all other nodes (the depth should not be smaller than the diameter of the graph). At the end of the network, the \READOUT\ function provides a pooling function to convert the representations to the appropriate output-level granularity (e.g. node-level or graph-level). Both the $\AGGREGATE$ and $\READOUT$ steps must be invariant to node permutations.  

\section{Theoretical Analysis}\label{sec:supp_theory}
\subsection{Controllability of the Representations from the Structure-Aware Attention}
\begin{theorem}
    % Assume that $f$ is a $\text{Lip}(f)$-Lipschitz mapping and the subgraph features extractor $\varphi$ is bounded by $C_{\varphi}$ on the space of subgraphs. For any pair of nodes $u$ and $v$ in two graphs $G=(V,E,\X)$ and $G'=(V',E',\X')$ with the same number of nodes $|V|=|V'|$, the distance between their representations after the structure-aware attention is controlled by the distance between their subgraph representations, through the following inequality:
    % \begin{equation}
    %     \|\text{SA-attn}(u)-\text{SA-attn}(v)\|\leq C_1 \left(\|h_u - h_v'\|+D(\H, \H')\right)+C_2 D(\X,\X'),
    % \end{equation}
    % where $h_w:=\varphi(w,G)$ denotes the subgraph representation at node $w$ for any $w\in V$ and $h_{w'}':=\varphi(w',G')$ similarly, and $\H=(h_w)_{w\in V}$ and $\H'=(h_{w'}')_{w'\in V'}$ denote the set of subgraph representations in $G$ and $G'$ respectively. $D$ is a matching metric between two multisets of representations with the same cardinality, defined as
    % \begin{equation*}
    %     D(\X,\X'):=\inf_{\pi\in\Pi(V,V')}\sup_{w\in V} \|x_w-x_{\pi(w)}'\|.
    % \end{equation*}
    Assume that $f$ is a Lipschitz mapping with the Lipschitz constant denoted by $\text{Lip}(f)$ and the structure extractor $\varphi$ is bounded by a constant $C_{\varphi}$ on the space of subgraphs. For any pair of nodes $v$ and $v'$ in two graphs $G=(V,E,\X)$ and $G'=(V',E',\X')$ with the same number of nodes $|V|=|V'|=n$, the distance between their representations after the structure-aware attention is bounded by:
    \begin{equation}
    % \begin{split}
         \|\text{SA-attn}(v)-\text{SA-attn}(v')\|\leq C_1 [\|h_v - h_{v'}'\| +D(\H, \H')] +C_2 D(\X,\X'),
    % \end{split}
    \end{equation}
    where $h_w:=\varphi(w,G)$ denotes the subgraph representation at node $w$ for any $w\in V$ and $h_{w'}':=\varphi(w',G')$ similarly, and $\H=(h_w)_{w\in V}$ and $\H'=(h_{w'}')_{w'\in V'}$ denote the multiset of subgraph representations in $G$ and $G'$ respectively. Denoting by $\Pi(V,V')$ the set of permutations between $V$ and $V'$, $D$ is a matching metric between two multisets of representations with the same cardinality, defined as
    \begin{equation*}
        D(\X,\X'):=\inf_{\pi\in\Pi(V,V')}\sup_{w\in V} \|x_w-x_{\pi(w)}'\|.
    \end{equation*}
    $C_1$ and $C_2$ are constants given by:
    \begin{equation*}
        C_1=\sqrt{\frac{2}{d_{out}}}n\text{Lip}(f) C_{\varphi} \|\Wq\|_{\infty} \|\Wk\|_{\infty}, ~~~
        C_2=\text{Lip}(f).
    \end{equation*}
\end{theorem}
\begin{proof}
    % Without loss of generality, we can assume that $G$ and $G'$ have the same number of nodes, that is $|V|=|V'|=n$, otherwise one can add virtual nodes to the smaller graph without adding any connectivity to the other nodes. 
    
    Let us denote by 
    \begin{equation*}
    \begin{aligned}
        z_v&=(\langle \Wq h_v, \Wk h_{w}\rangle)_{w\in V}\in\real^n, \\
        z_{v'}'&=(\langle \Wq h_{v'}', \Wk h_{w'}'\rangle)_{w'\in V'}\in\real^n,
    \end{aligned}
    \end{equation*}
    and by $\softmax(z)\in\real^n$ for any $z\in\real^n$ with its $i$-th coefficient
    \begin{equation*}
        \softmax(z)_i=\frac{\exp(z_i/\sqrt{d_{out}})}{\sum_{j=1}^n \exp(z_j/\sqrt{d_{out}})}.
    \end{equation*}
    Then, we have
    \begin{equation*}
        \begin{aligned}
            & \|\text{SA-Attn}(v)-\text{SA-Attn}(v')\| \\
            =& \left\| \sum_{w\in V} \softmax(z_v)_w f(x_w) -\sum_{w'\in V'} \softmax(z_{v'}')_{w'} f(x_{w'}')\right\| \\
            =& \left\| \sum_{w\in V} (\softmax(z_v)_w -\softmax(z_{v'}')_{\pi(w)}) f(x_w)+\sum_{w\in V}\softmax(z_{v'}')_{\pi(w)} f(x_w) -\sum_{w'\in V'} \softmax(z_{v'}')_{w'} (f(x_{w'}'))\right\| \\
            \leq & \left\| \sum_{w\in V} (\softmax(z_v)_w -\softmax(z_{v'}')_{\pi(w)}) f(x_w)\right\| + \left\| \sum_{w'\in V'}\softmax(z_{v'}')_{w'} (f(x_{\pi^{-1}(w')}) - f(x_{w'}'))\right\| 
            % &\leq \|\softmax(z_u)-\softmax(z_v')\| \sqrt{\sum_{i=1}^n \|f(x_i)\|^2} + \sum_{j=1}^n \softmax(z_v')_j \left\|f(x_i)-f(x_j')\right\| \\
            % &=
        \end{aligned}
    \end{equation*}
    where $\pi:V\to V'$ is an arbitrary permutation and we used the triangle inequality. Now we need to bound the two terms respectively. We first bound the second term:
    \begin{equation*}
        \begin{aligned}
            \left\| \sum_{w'\in V'}\softmax(z_{v'}')_{w'} (f(x_{\pi^{-1}(w')}) - f(x_{w'}'))\right\| &
            \leq \sum_{w'\in V'}\softmax(z_{v'}')_{w'} \left\|f(x_{\pi^{-1}(w')}) - f(x_{w'}')\right\| \\
            &\leq \sum_{w'\in V'}\softmax(z_{v'}')_{w'} \text{Lip}(f)\|x_{\pi^{-1}(w')} - x_{w'}'\| \\
            &= \text{Lip}(f)\sum_{w'\in V'}\softmax(z_{v'}')_{w'}\|x_{\pi^{-1}(w')} - x_{w'}'\| \\
            &\leq \text{Lip}(f)\sup_{w'\in V'}\|x_{\pi^{-1}(w')} - x_{w'}'\| \\
            &= \text{Lip}(f)\sup_{w\in V}\|x_{w} - x_{\pi(w)}'\| 
            % &= \text{Lip}(f)\sup_{w\in V}\|x_{w} - x_{\pi(w)}'\| \\
            % &\leq \text{Lip}(f)\sup_{w\in V, w'\in V'}\|x_{w} - x_{w'}'\|,
            % &=  \text{Lip}(f)\left\|x-x_j'\right\| 
        \end{aligned}
    \end{equation*}
    where the first inequality is a triangle inequality, the second inequality uses the Lipschitzness of $f$. And for the first term, we can upper-bound it by
    \begin{equation*}
        \begin{aligned}
            & \left\| \sum_{w\in V} (\softmax(z_{v})_w -\softmax(z_{v'}')_{\pi(w)}) f(x_w)\right\| \\
            \leq & \|\softmax(z_v)-\softmax((z_{v'}')_{\pi})\| \sqrt{\sum_{w\in V} \|f(x_w)\|^2}  \\
            \leq & \frac{1}{\sqrt{d_{out}}} \| z_v - (z_{v'}')_{\pi}\| \sqrt{n} \text{Lip}(f),
        \end{aligned}
    \end{equation*}
    where by abuse of notation, $(z)_{\pi}\in\real^n$ denotes the vector whose $w$-th entry is $z_{\pi(w)}$ for any $z\in\real^n$. The first inequality comes from a simple matrix norm inequality, and the second inequality uses the fact that $\softmax$ function is $1/\sqrt{d_{out}}$-Lipschitz (see \eg \citet{gao2017properties}). Then, we have
    \begin{equation*}
        \begin{aligned}
            \| z_v - (z_{v'}')_{\pi}) \|^2&=\sum_{w\in V}\left(\langle \Wq h_v, \Wk h_w\rangle-\langle \Wq h_{v'}', \Wk h_{\pi(w)}'\rangle\right)^2 \\
            & = \sum_{w\in V}\left(\langle \Wq h_v, \Wk (h_w-h_{\pi(w)}')\rangle+\langle \Wq(h_v- h_{v'}'), \Wk h_{\pi(w)}'\rangle\right)^2 \\
            & \leq 2\sum_{w\in V}\left(\langle \Wq h_v, \Wk (h_w-h_{\pi(w)}')\rangle^2+\langle \Wq(h_v- h_{v'}'), \Wk h_{\pi(w)}'\rangle^2\right) \\
            & \leq 2\sum_{w\in V}\left(\|\Wq h_v\|^2\|\Wk (h_w-h_{\pi(w)}')\|^2+\|\Wq(h_v- h_{v'}')\|^2\| \Wk h_{\pi(w)}'\|^2\right) \\
            & \leq 2\sum_{w\in V}\left(C_{\varphi}^2\|\Wq\|_{\infty}^2 \|\Wk\|_{\infty}^2\| h_w-h_{\pi(w)}'\|^2+\|\Wq\|_{\infty}^2\|h_v- h_{v'}'\|^2C_{\varphi}^2\|\Wk\|_{\infty}^2\right) \\
            & \leq 2n C_{\varphi}^2\|\Wq\|_{\infty}^2 \|\Wk\|_{\infty}^2 \left(\|h_v- h_{v'}'\|^2+\sup_{w\in V}\| h_w-h_{\pi(w)}'\|^2\right),
            % & \leq 2n C_{\varphi}^2\|\Wq\|_{\infty}^2 \|\Wk\|_{\infty}^2 \left(\|h_u- h_v'\|^2+\sup_{w\in V, w'\in V'}\| h_w-h_{w'}'\|^2\right).
        \end{aligned}
    \end{equation*}
    where the first inequality comes from $(a+b)^2\leq 2(a^2+b^2)$, the second one uses the Cauchy-Schwarz inequality and the third one uses the definition of spectral norm and the bound of the structure extractor function. Then, we obtain the following inequality
    \begin{equation*}
    \begin{aligned}
        & \left\| \sum_{w\in V} (\softmax(z_v)_w -\softmax(z_{v'}')_{\pi(w)}) f(x_w)\right\| \\ 
        \leq & \sqrt{\frac{2}{d_{out}}}n\text{Lip}(f) C_{\varphi} \|\Wq\|_{\infty} \|\Wk\|_{\infty}\left(\|h_v- h_{v'}'\|+\sup_{w\in V}\| h_w-h_{\pi(w)}'\|\right)
    \end{aligned}
    \end{equation*}
    By combining the upper bounds of the first and the second term, we obtain an upper bound for the distance between the structure-aware attention representations:
    \begin{equation*}
        \|\text{SA-attn}(v)-\text{SA-attn}(v')\| \leq C_1 \left(\|h_v- h_{v'}'\|+\sup_{w\in V}\| h_w-h_{\pi(w)}'\|\right) + C_2 \sup_{w\in V}\|x_{w} - x_{\pi(w)}'\|,
    \end{equation*}
    for any permutation $\pi\in \Pi(V,V')$, where 
    \begin{equation*}
        \begin{aligned}
            C_1&=\sqrt{\frac{2}{d_{out}}}n\text{Lip}(f) C_{\varphi} \|\Wq\|_{\infty} \|\Wk\|_{\infty} \\
            C_2&=\text{Lip}(f).
        \end{aligned}
    \end{equation*}
    Finally, by taking the infimum over the set of permutations, we obtain
    the inequality in the theorem.
    % \begin{equation*}
    %     \begin{aligned}
    %         \|\text{SA-Attn}(u)-\text{SA-Attn}(v)\| &= \left\| \sum_{v'\in V} \frac{\kappa_{\exp}(h_u, h_{v'})}{\sum_{w\in V} \kappa_{\exp}(h_u,h_w)}f(x_{v'}) - \sum_{v'\in V} \frac{\kappa_{\exp}(h_v, h_{v'})}{\sum_{w\in V} \kappa_{\exp}(h_v,h_w)}f(x_{v'}) \right\| \\
    %         &=\left\| \sum_{v'\in V} \left(\frac{\kappa_{\exp}(h_u, h_{v'})}{\sum_{w\in V} \kappa_{\exp}(h_u,h_w)} -\frac{\kappa_{\exp}(h_v, h_{v'})}{\sum_{w\in V} \kappa_{\exp}(h_v,h_w)}\right)f(x_{v'})\right\| \\
    %         &\leq \sum_{v'\in V}\left|\frac{\kappa_{\exp}(h_u, h_{v'})}{\sum_{w\in V} \kappa_{\exp}(h_u,h_w)} -\frac{\kappa_{\exp}(h_v, h_{v'})}{\sum_{w\in V} \kappa_{\exp}(h_v,h_w)}\right| \left\| f(x_{v'})\right\| \\
    %     \end{aligned}
    % \end{equation*}
\end{proof}

\subsection{Expressivity Analysis}
Here, we assume that $f$ can be any continuous mapping and it is approximated by an MLP network through the universal approximation theorem~\citep{hornik1991approximation} in practice.
\begin{theorem}
    % Assume that the space of node attributes is countable. For any pair of nodes $u$ and $v$ in two graphs $G=(V,E,\X)$ and $G'=(V',E',\X')$, assume that there exists a node $v'$ in $V$ such that $x_v$ is different from all other nodes and a node $w$ in $V$ that its subgraph representation $\varphi(w,G)$ is different from all other subgraph representations. Then, there exists a set of parameters and a mapping $f:\Xcal\to \real^{d_{out}}$ such that their representations after the structure-aware attention are different, i.e.\ $\text{SA-attn}(u)\neq \text{SA-attn}(v)$, if their subgraph representations are different, i.e.\ $\varphi(u,G)\neq \varphi(v, G')$.
    Assume that the space of node attributes $\Xcal$ is countable. For any pair of nodes $v$ and $v'$ in two graphs $G=(V,E,\X)$ and $G'=(V',E',\X')$, assume that there exists a node $u_1$ in $V$ such that $x_{u_1}\neq x_{w}$ for any $w\in V$ and a node $u_2$ in $V$ such that its subgraph representation $\varphi(u_2,G)\neq \varphi(w,G)$ for any $w\in V$. Then, there exists a set of parameters and a mapping $f:\Xcal\to \real^{d_{out}}$ such that their representations after the structure-aware attention are different, \ie $\text{SA-attn}(v)\neq \text{SA-attn}(v')$, if their subgraph representations are different, \ie $\varphi(v,G)\neq \varphi(v', G')$.
\end{theorem}
\begin{proof}
    This theorem amounts to showing the injectivity of the original dot-product attention with respect to the query, that is to show
    \begin{equation*}
        \text{Attn}(h_v,x_v, G)=\sum_{u\in V} \frac{\kappa_{\exp}(h_v, h_{u})}{\sum_{w\in V} \kappa_{\exp}(h_v,h_w)}f(x_{u})
    \end{equation*}
    is injective in $h_v$, where
    \begin{equation}
        \kappa_{\exp}(h, h'):=\exp\left(\langle\Wq h+b_Q,\Wk h'+b_K\rangle/\sqrt{d_{out}}\right).
    \end{equation}
    Here we consider the offset terms that were omitted in Eq.~\eqref{eq:Attention}.
    Let us prove the contrapositive of the theorem. We assume that $\text{Attn}(h_v,x_v,G)= \text{Attn}(h_{v'}',x_{v'}',G')$ for any set of parameters and any mapping $f$ and want to show that $h_v=h_{v'}'$. 
    
    Without loss of generality, we assume that $G$ and $G'$ have the same number of nodes, that is $|V|=|V'|=n$. Otherwise, one can easily add some virtual isolated nodes to the smaller graph. Now if we take $\Wq=\Wk=0$, all the softmax coefficients will be identical and we have
    \begin{equation*}
        \sum_{w\in V} f(x_{w})=\sum_{w'\in V'} f(x_{w'}').
    \end{equation*}
    Thus, by Lemma 5 of \citet{xu2018how}, there exists a mapping $f$ such that the multisets $X$ and $X'$ are identical. 
    
    As a consequence, we can re-enumerate the nodes in two graphs by a sequence $V$ (by abuse of notation, we keep using $V$ here) such that $x_u=x_u'$ for any $u\in V$. Then, we can rewrite the equality $\text{Attn}(h_v,x_v,G)= \text{Attn}(h_{v'}',x_{v'}',G')$ as
    \begin{equation*}
        \sum_{u\in V} \left(\frac{\kappa_{\exp}(h_v, h_{u})}{\sum_{w\in V} \kappa_{\exp}(h_v,h_w)}-\frac{\kappa_{\exp}(h_{v'}', h_{u}')}{\sum_{w\in V} \kappa_{\exp}(h_{v'}',h_w')}\right)f(x_{u})=0.
    \end{equation*}
    Now since there exists a node $u_1$ in $V$ such that its attributes are different from all other nodes, \ie $x_{u_1}\neq x_{w}$ for any $w\in V$, we can find a mapping $f$ such that $f(x_{u_1})$ is not in the span of $(f(x_w))_{w\in V,w\neq u_1}$. Then, by their independence we have
    \begin{equation*}
        \frac{\kappa_{\exp}(h_v, h_{u_1})}{\sum_{w\in V} \kappa_{\exp}(h_v,h_w)}=\frac{\kappa_{\exp}(h_{v'}', h_{u_1}')}{\sum_{w\in V} \kappa_{\exp}(h_{v'}',h_w')},
    \end{equation*}
    for any $\Wq$, $\Wk$, $b_Q$ and $b_K$.
    
    On the one hand, if we take $\Wq=0$, we have for any $\Wk$, $b_Q$ and $b_K$ that
    \begin{equation*}
        \frac{\exp{(\langle b_Q, \Wk h_{u_1}+b_K\rangle/\sqrt{d_{out}})}}{\sum_{w\in V} \exp{(\langle b_Q, \Wk h_w+b_K\rangle/\sqrt{d_{out}})}}=\frac{\exp{(\langle b_Q, \Wk h_{u_1}'+b_K\rangle/\sqrt{d_{out}})}}{\sum_{w\in V} \exp{(\langle b_Q, \Wk h_w'+b_K\rangle/\sqrt{d_{out}})}}.
    \end{equation*}
    On the other hand if we take $b_Q=0$ we have for any $\Wq$, $\Wk$ and $b_K$ that
    \begin{equation*}
    \begin{aligned}
        \frac{\exp{(\langle \Wq h_v, \Wk h_{u_1}+b_K\rangle/\sqrt{d_{out}})}}{\sum_{w\in V} \exp{(\langle \Wq h_v, \Wk h_w+b_K\rangle/\sqrt{d_{out}})}}&=\frac{\exp{(\langle \Wq h_{v'}', \Wk h_{u_1}'+b_K\rangle/\sqrt{d_{out}})}}{\sum_{w\in V} \exp{(\langle \Wq h_{v'}', \Wk h_w'+b_K\rangle/\sqrt{d_{out}})}} \\
        &=\frac{\exp{(\langle \Wq h_{v'}', \Wk h_{u_1}+b_K\rangle/\sqrt{d_{out}})}}{\sum_{w\in V} \exp{(\langle \Wq h_{v'}', \Wk h_w+b_K\rangle/\sqrt{d_{out}})}},
    \end{aligned}
    \end{equation*}
    where the second equality is obtained by replacing $b_Q$ with $\Wq h_{v'}'$ in the above equality. Then, we can rewrite the above equality as below:
    \begin{equation*}
        \sum_{w\in V}\exp{\left(\frac{\langle \Wq h_v, \Wk (h_w-h_{u_1})\rangle}{\sqrt{d_{out}}}\right)}=\sum_{w\in V}\exp{\left(\frac{\langle \Wq h_{v'}', \Wk (h_w-h_{u_1})\rangle}{\sqrt{d_{out}}}\right)}.
    \end{equation*}
    If we denote by $\phi:\real^{d_{out}}\to \Hcal$ the feature mapping associated with the dot product kernel $\kappa_{\text{exp}}(t,t')=\exp(\langle t,t'\rangle/\sqrt{d_{out}})$ and $\Hcal$ the correspond reproducing kernel Hilbert space, we then have for any $\Wq$ and $\Wk$ that
    \begin{equation*}
        \left\langle \phi(\Wq h_v) - \phi(\Wq h_{v'}'), \sum_{w\in V}\phi(\Wk(h_w - h_{u_1}))\right\rangle_{\Hcal}=0.
    \end{equation*}
    Since by assumption there exists a $u_2\in V$ such that $h_{u_2}-h_{u_1}\neq 0$ and $\kappa_{\text{exp}}$ is a universal kernel~\citep{micchelli2006universal}, $\Wk\mapsto\phi(\Wk(h_{u_2}-h_{u_1}))$ is dense in $\Hcal$ and we have $\phi(\Wq h_v)=\phi(\Wq h_{v'}')$. We can then conclude, by the injectivity of $\phi$, that
    \begin{equation*}
        \Wq h_v=\Wq h_{v'}',
    \end{equation*}
    for any $\Wq$, and thus $h_v=h_{v'}'$. Now by taking $h_v=\varphi(v, G)$ and $h_{v'}'=\varphi(v',G')$, we obtain the theorem.
\end{proof}

\section{Experimental Details and Additional Results}\label{sec:supp_experiments}
In this section, we provide implementation details and additional experimental results.  %To preserve anonymity in the review process, we will censor the GitHub link to our code~(\censor{https:/www.github.com}), but will make it publicly available upon publication.  

\subsection{Computation Details}
All experiments were performed on a shared GPU cluster equipped with GTX1080, GTX1080TI, GTX2080TI and TITAN RTX. About 20 of these GPUs were used simultaneously, and the total computational cost of this research project was about 1k GPU hours.

\subsection{Datasets Description}
We provide details of the datasets used in our experiments, including ZINC~\citep{ZINC}, CLUSTER~\citep{dwivedi2020benchmarkgnns}, PATTERN~\citep{dwivedi2020benchmarkgnns}, OGBG-PPA~\citep{hu2020open} and OGBG-CODE2~\citep{hu2020open}. For each dataset, we follow their respective training protocols and use the standard train/validation/test splits and evaluation metrics.

\paragraph{ZINC.} The ZINC dataset is a graph regression dataset comprised of molecules, where the task is to predict constrained solubility. Like \citet{dwivedi2020benchmarkgnns}, we use the subset of 12K molecules and follow their same splits.

\paragraph{PATTERN and CLUSTER.} PATTERN and CLUSTER \citet{dwivedi2020benchmarkgnns} are synthetic datasets that were created using Stochastic Block Models \citep{PatternCluster2018}. The goal for both datasets is node classification, with PATTERN focused on detecting a given pattern in the dataset, and with CLUSTER focused on identifying communities within the graphs. For PATTERN, the binary class label corresponds to whether a node is part of the predefined pattern or not; for CLUSTER, the multi-class label indicates membership in a community. We use the splits as is used in \citet{dwivedi2020benchmarkgnns}.

\paragraph{OGBG-PPA.} PPA \citep{hu2020open} is comprised of protein-protein association networks where the goal is to correctly classify the network into one of 37 classes representing the category of species the network is from. Nodes represent proteins and edges represent associations between proteins. Edge attributes represent information relative to the association, such as co-expression. We use the standard splits provided by~\citet{hu2020open}.

\paragraph{OGBG-CODE2.} CODE2 \cite{hu2020open} is a dataset containing source code from the Python programming language. It is made up of Abstract Syntax Trees where the task is to correctly classify the sub-tokens that comprise the method name. We use the standard splits provided by~\citet{hu2020open}.

\subsection{Hyperparameter Choices and Reproducibility}\label{sec:supp_hyperparameter}
% optimization and hyperparameters
\paragraph{Hyperparameter choice.}
In general, we perform a very limited hyperparameter search to produce the results in Table~\ref{tab:comparison_to_SOTA} and Table~\ref{tab:comparison_to_ogb_SOTA}. The hyperparameters for training SAT models on different datasets are summarized in Table~\ref{tab:supp_hyperparameter}, where only the dropout rate and the size of the subgraph $k$ are tuned ($k$ $\in \{1,2,3,4\}$). We use fixed RWPE~\citep{dwivedi2022graph} with SAT on ZINC, PATTERN and CLUSTER. In all experiments, we use the validation set to select the dropout rate and the size of the subtree or subgraph $k$ $\in \{1,2,3,4\}$. All other hyperparameters are fixed for simplicity, including setting the readout method to mean pooling. We did not use RWPE on OGBG-PPA and OGBG-CODE2 as we observed very little performance improvement. Note that we only use $k=1$ for the $k$-subgraph SAT models on CLUSTER and PATTERN due to its large memory requirement, which already leads to performance boost compared to the $k$-subtree SAT using a larger $k$. Reported results are the average over 4 seeds on ZINC, PATTERN and CLUSTER, as is done in~\citet{dwivedi2020benchmarkgnns}, and averaged over 10 seeds on OGBG-PPA and OGBG-CODE2.

\begin{table}[]
    \centering
    \caption{Hyperparameters for SAT models trained on different datasets. RWPE-$p$ indicates using $p$ steps in the random walk positional encoding, which results in a $p$-dimensional vector as the positional representation for each node.}
    \label{tab:supp_hyperparameter}
    \begin{tabular}{lccccc}
        \toprule
        %  & \multicolumn{3}{c}{\textsc{Dataset}} \\
         Hyperparameter & \textbf{ZINC} & \textbf{CLUSTER} & \textbf{PATTERN} & \textbf{OGBG-PPA} & \textbf{OGBG-CODE2} \\
         \midrule
         \#Layers & 6 & 16 & 6 & 3 & 4 \\
         Hidden dimensions & 64 & 48 & 64 & 128 & 256 \\
         FFN hidden dimensions & \multicolumn{5}{c}{$2\times$Hidden dimensions} \\
         \#Attention heads & 8 & 8 & 8 & 8 & $\{\mathbf{4},8\}$ \\
         Dropout & \multicolumn{5}{c}{$\{0.0, 0.1, 0.2, 0.3, 0.4\}$} \\
         Size of subgraphs $k$ & \multicolumn{5}{c}{$\{1, 2, 3, 4\}$} \\
         Readout method & mean & None & None & mean & mean \\
         Absolute PE & RWPE-20 & RWPE-3 & RWPE-7 & None & None \\ \midrule
         Learning rate & 0.001 & 0.0005 & 0.0003 & 0.0003 & 0.0001 \\
         Batch size & 128 & 32 & 32 & 32 & 32 \\
         \#Epochs & 2000 & 200 & 200 & 200 & 30 \\
         Warm-up steps & 5000 & 5000 & 5000 & 10 epochs & 2 epochs \\
         Weight decay & 1e-5 & 1e-4 & 1e-4 & 1e-4 & 1e-6 \\
         \bottomrule
    \end{tabular}
\end{table}

\paragraph{Optimization.}
All our models are trained with the AdamW optimizer~\citep{loshchilov2018decoupled} with a standard warm-up strategy suggested for Transformers in~\citet{vaswani2017attention}. We use either the L1 loss or the cross-entropy loss depending on whether the task is regression or classification. The learning rate scheduler proposed in the Transformer is used on the ZINC, PATTERN and CLUSTER datasets and a cosine scheduler~\citep{loshchilov2016sgdr} is used on the larger OGBG-PPA and OGBG-CODE2 datasets.

\paragraph{Number of parameters and computation time.}
In Table~\ref{tab:supp_number_parameter}, we report the number of parameters and the training time per epoch for SAT with $k$-subtree GNN extractors using the hyperparameters selected from Table~\ref{tab:supp_hyperparameter}. Note that the number of parameters used in our SAT on OGB datasets is smaller than most of the state-of-art methods.

\begin{table}[]
    \centering
    \caption{Number of parameters and training time per epoch for $k$-subtree SAT models using the hyperparameters in Table~\ref{tab:supp_hyperparameter}. Various GNNs are used as the base GNN in SAT.}
    \label{tab:supp_number_parameter}
    \begin{tabular}{lccccc}
        \toprule
        %  & \multicolumn{3}{c}{\textsc{Dataset}} \\
          & \textbf{ZINC} & \textbf{CLUSTER} & \textbf{PATTERN} & \textbf{OGBG-PPA} & \textbf{OGBG-CODE2} \\
         \midrule
         Base GNN & \multicolumn{5}{c}{\#Parameters} \\
         GCN & 421k & 571k & 380k & 766k & 14,030k \\
         GIN & 495k & 684k & 455k & 866k & 14,554k \\
         PNA & 523k & 741k & 493k & 1,088k & 15,734k \\ 
         \midrule
         Base GNN & \multicolumn{5}{c}{GPU time on a single TITAN RTX/epoch} \\ 
         GCN & 6s & 142s & 40s & 308s & 40min \\
         GIN & 6s & 144s & 62s & 310s & 40min \\
         PNA & 9s & 178s & 90s & 660s & 55min \\
         \bottomrule
    \end{tabular}
\end{table}

\subsection{Additional Results}
We provide additional experimental results on ZINC, OGBG-PPA and OGBG-CODE2.

\subsubsection{Additional Results on ZINC}
We report a more thorough comparison of SAT instances using different structure extractors and different readout methods in Table~\ref{tab:supp_zinc}. We find that SAT models with PNA consistently outperform other GNNs. Additionally, the readout methods have very little impact on the prediction performance.

\begin{table}
\centering
  \caption{Test MAE for SAT models using different structure extractors and readout methods on the ZINC dataset.}
  \label{tab:supp_zinc}
  \resizebox{\textwidth}{!}{
    % \begin{small}
    % \setlength{\tabcolsep}{0.50em}
      \begin{sc}
        %\resizebox{\textwidth}{!}{%
          \begin{tabular}{ll|ccc|ccc}
            \toprule
            % \textbf{ZINC} \\
            & & \multicolumn{3}{c}{\textbf{w/o edge attributes}} & \multicolumn{3}{c}{\textbf{w/ edge attributes}} \\ \midrule
            & Base GNN & Mean & Sum & Cls & Mean & Sum & Cls \\ 
            \midrule
            \multirow{4}{*}{k-subtree SAT} 
            & GCN & 0.174$\pm$0.009 & 0.170$\pm$0.010 & 0.167$\pm$9.005 & 0.127$\pm$0.010 & 0.117$\pm$0.008 & 0.115$\pm$0.007 \\ %& 0.102$\pm$0.005 & 0.102$\pm$0.003 & \textbf{0.098$\pm$0.008} \\
            & GIN & 0.166$\pm$0.007 & 0.162$\pm$0.010 & 0.157$\pm$0.002 & 0.115$\pm$0.005 & 0.112$\pm$0.008 & 0.104$\pm$0.003 \\
            & GraphSAGE & 0.164$\pm$0.004 & 0.165$\pm$0.008 & 0.156$\pm$0.005 & - & - & - \\
            & PNA & 0.147$\pm$0.001 & 0.142$\pm$0.008 & \textbf{0.135$\pm$0.004} & 0.102$\pm$0.005 & 0.102$\pm$0.003 & \textbf{0.098$\pm$0.008} \\ \midrule
            \multirow{4}{*}{k-subgraph SAT} 
            & GCN & 0.184$\pm$0.002 & 0.186$\pm$0.007 & 0.184$\pm$0.007 & 0.114$\pm$0.005 & 0.103$\pm$0.002 & 0.103$\pm$0.008 \\ %0.094$\pm$0.008 & \textbf{0.089$\pm$0.002} & 0.093$\pm$0.009 \\
            & GIN & 0.162$\pm$0.013 & 0.158$\pm$0.007 & 0.162$\pm$0.005 & 0.095$\pm$0.002 & 0.097$\pm$0.002 & 0.098$\pm$0.010 \\
            & GraphSAGE & 0.168$\pm$0.005 & 0.165$\pm$0.005 & 0.169$\pm$0.005 & - & - & -\\
            & PNA & 0.131$\pm$0.002 & 0.129$\pm$0.003 & \textbf{0.128$\pm$0.004} & 0.094$\pm$0.008 & \textbf{0.089$\pm$0.002} & 0.093$\pm$0.009 \\
            \bottomrule
        \end{tabular}
        %}
      \end{sc}
    % \end{small}
    }
\end{table}

\subsubsection{Additional Results on OGBG-PPA}
Table~\ref{tab:supp_ppa} summarizes the results for $k$-subtree SAT with different GNNs compared to state-of-the-art methods on OGBG-PPA. All the results are computed from 10 runs using different random seeds.

\begin{table}
  \centering
  \caption{Comparison of SAT and SOTA methods on the OGBG-PPA dataset. All results are computed from 10 different runs.}
  \label{tab:supp_ppa}
  \begin{sc}
        \begin{tabular}{lccc}
        \toprule
        {}& \multicolumn{2}{c}{\textbf{OGBG-PPA}} \\
        Method & Test accuracy & Validation accuracy \\
        \midrule
        GCN & 0.6839$\pm$0.0084 & 0.6497$\pm$0.0034 \\
        GCN-Virtual Node & 0.6857$\pm$0.0061 & 0.6511$\pm$0.0048	\\
        GIN & 0.6892$\pm$0.0100 & 0.6562$\pm$0.0107 \\
        GIN-Virtual Node & 0.7037$\pm$0.0107 & 0.6678$\pm$0.0105 \\
        \midrule
        Transformer & 0.6454$\pm$0.0033 & 0.6221$\pm$0.0039 \\
        \midrule
        k-subtree SAT-GCN & 0.7483$\pm$0.0048 & \textbf{0.7072$\pm$0.0030} \\
        k-subtree SAT-GIN & 0.7306$\pm$0.0076 & 0.6928$\pm$0.0058 \\
        k-subtree SAT-PNA & \textbf{0.7522$\pm$0.0056} & 0.7025$\pm$0.0064 \\
        \bottomrule
        \end{tabular}
    \end{sc}
\end{table}

\subsubsection{Additional Results on OGBG-CODE2}
Table~\ref{tab:supp_code2} summarizes the results for $k$-subtree SAT with different GNNs compared to state-of-the-art methods on OGBG-CODE2. All the results are computed from 10 runs using different random seeds.

\begin{table}
  \caption{Comparison of SAT and SOTA methods on the OGBG-CODE2 dataset. All results are computed from 10 different runs.}
  \label{tab:supp_code2}
  \centering
  \begin{sc}
        \begin{tabular}{lccc}
        \toprule
        {}& \multicolumn{2}{c}{\textbf{OGBG-CODE2}} \\
        Method & Test F1 score & Validation F1 score \\
        \midrule
        GCN & 0.1507$\pm$0.0018 & 0.1399$\pm$0.0017 \\
        GCN-Virtual Node & 0.1581$\pm$0.0026 & 0.1461$\pm$0.0013	\\
        GIN & 0.1495$\pm$0.0023 & 0.1376$\pm$0.0016 \\
        GIN-Virtual Node & 0.1581$\pm$0.0026 & 0.1439$\pm$0.0020 \\
        \midrule
        Transformer & 0.1670$\pm$0.0015 &  0.1546$\pm$0.0018 \\
        GraphTrans & 0.1830$\pm$0.0024 & 0.1661$\pm$0.0012 \\
        \midrule
        k-subtree SAT-GCN & 0.1934$\pm$0.0020 & \textbf{0.1777$\pm$0.0011} \\
        k-subtree SAT-GIN & 0.1910$\pm$0.0023 & 0.1748$\pm$0.0016 \\
        k-subtree SAT-PNA & \textbf{0.1937$\pm$0.0028} & 0.1773$\pm$0.0023 \\
        \bottomrule
        \end{tabular}
    \end{sc}
\end{table}

\begin{comment}
\subsection{Additional Experiments on Positional Encoding}

\leo{We now provide additional experiments to showcase the role of the positional encoding, by including a comparison of when the positional encoding is random. To generate random encodings, we took the existing positional encodings of a graph and randomly permuted them (RW-perm, LAP-perm), as well as generated a random positional encoding based on white noise (Random). We show the results in the figure below. We include the results shown in the main paper (None, LAP, RW). The main paper only compares the LAP and RW positional encodings to no positional encoding (None), as ``No positional encoding" best isolates the effect of positional encoding, whereas a random positional encoding would remove the positional encoding, but also add noise to the model. 
}
\begin{figure}[htbp]
    \centering
    %\vspace{-.3cm}
    \input{figures/pos_enc_rebuttal}
    \caption{A comparison of the effect of a positional encoding (RW, LAP) to no positional encoding (None), a white noise positional encoding (Random), and random permutations of the RW and LAP positional encodings (RW-perm, LAP-perm.}
    \vspace{-.7cm}
\end{figure}
\end{comment}

\newpage

\section{Model Interpretation}\label{sec:supp_interpretation}
In this section, we provide implementation details about the model visualization.

\subsection{Dataset and Training Details}
We use the Mutagenicity dataset~\citep{KKMMN2016}, consisting of 4337 molecular graphs labeled based on their mutagenic effect. We randomly split the dataset into train/val/test sets in a stratified way with a proportion of 80/10/10. We first train a two-layer vanilla Transformer model using RWPE. The hidden dimension and the number of heads are fixed to 64 and 8 respectively. The CLS pooling as described in Section~\ref{sec:pooling} is chosen as the readout method for visualization purpose. We also train a $k$-subtree SAT using exactly the same hyperparameter setting except that \emph{it does not use any absolute positional encoding}. $k$ is fixed to 2. For both models, we use the AdamW optimizer and the optimization strategy described in Section~\ref{sec:supp_hyperparameter}. We train enough epochs until both models converge. While the classic Transformer with RWPE achieves a test accuracy of 78\%, the $k$-subtree SAT achieves a 82\% test accuracy.

\subsection{Additional Results}

\begin{figure}[htbp]
    \includegraphics[width=.4\textwidth]{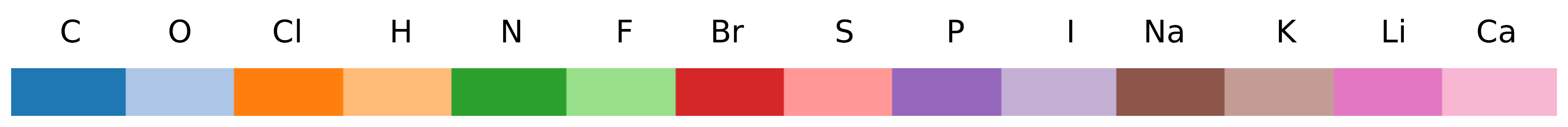} \\
    \begin{center}
    % {\centering
    \includegraphics[width=.48\textwidth]{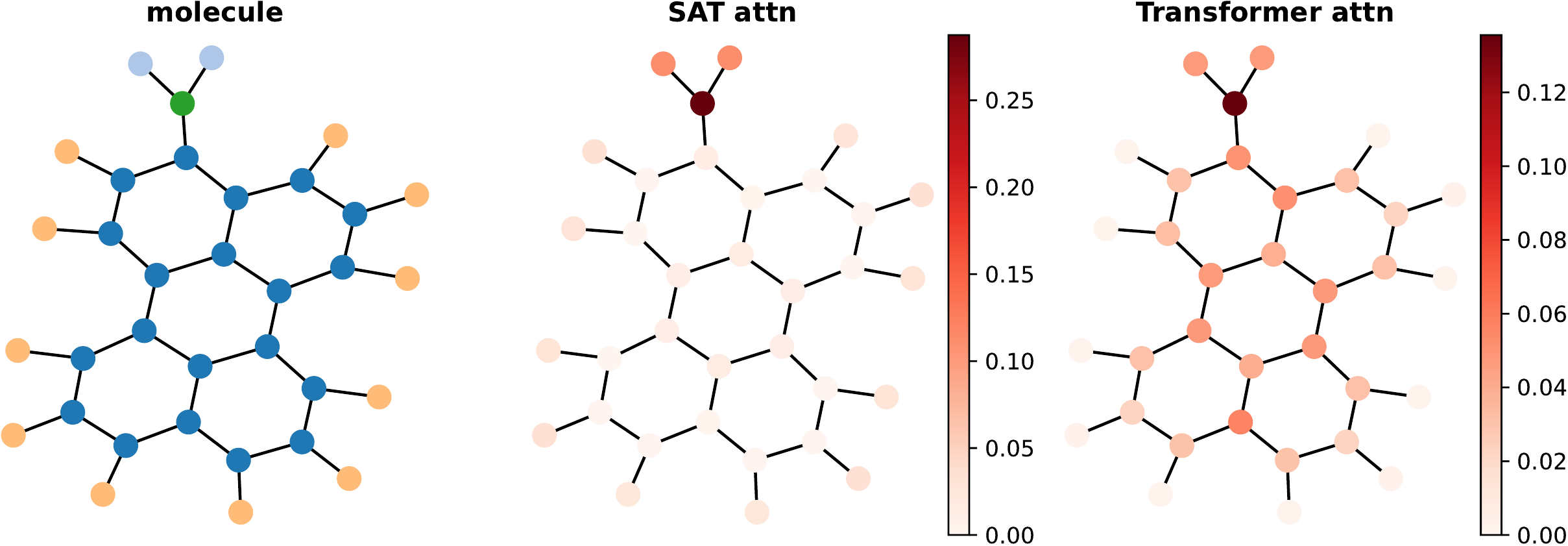}
    \includegraphics[width=.48\textwidth]{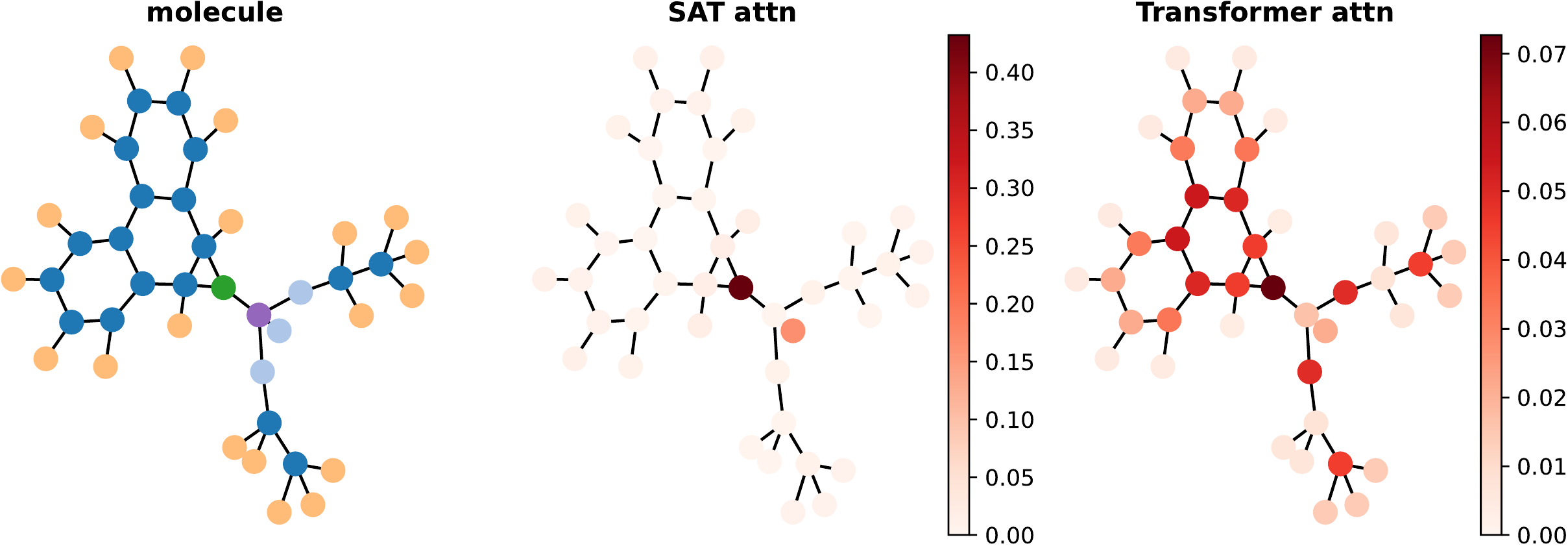} \\
    \includegraphics[width=.48\textwidth]{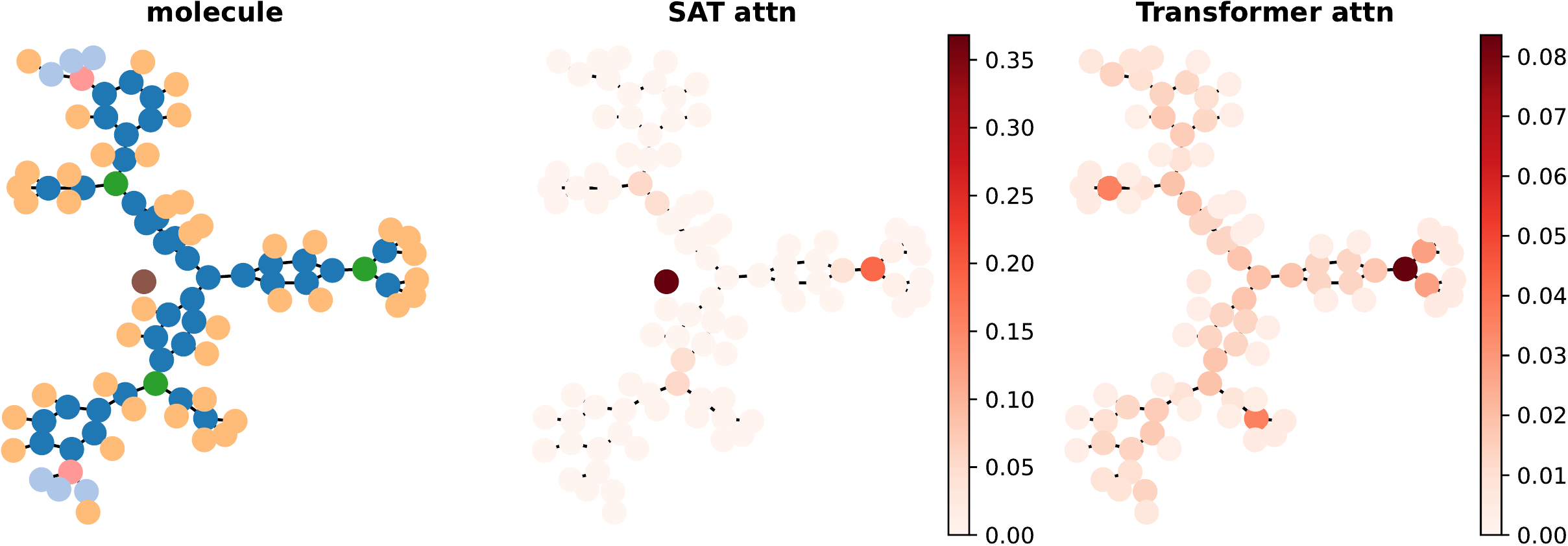}
    \includegraphics[width=.48\textwidth]{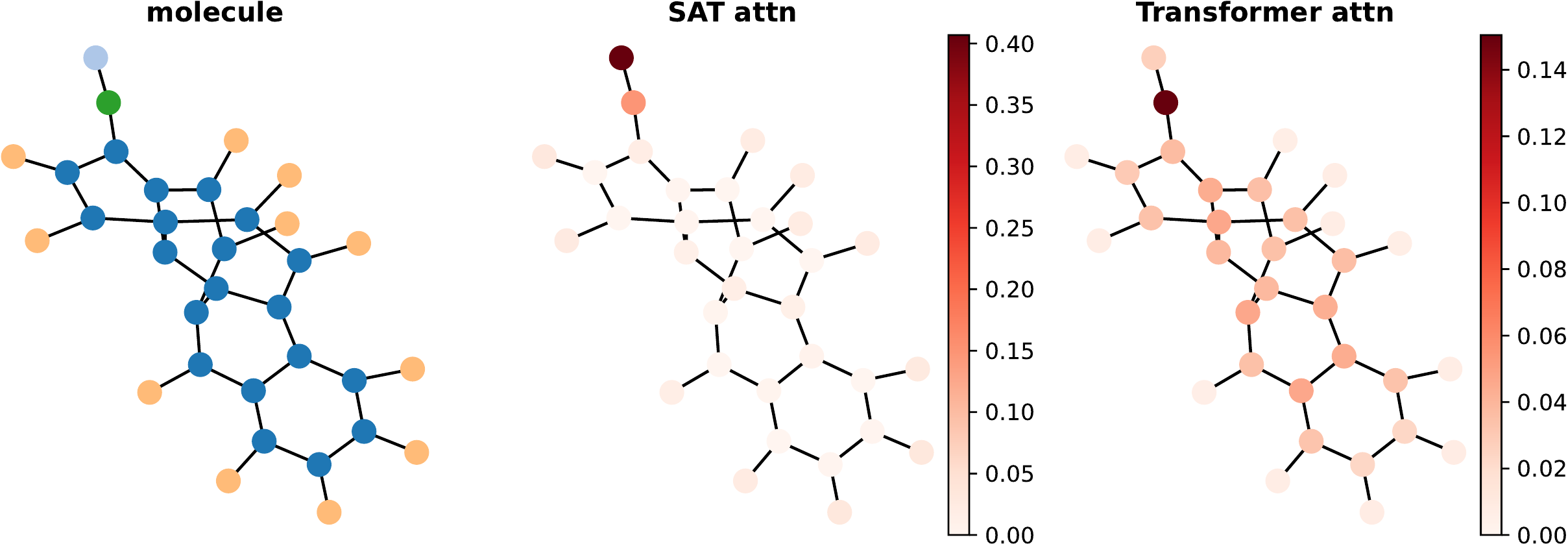} \\
    \includegraphics[width=.48\textwidth]{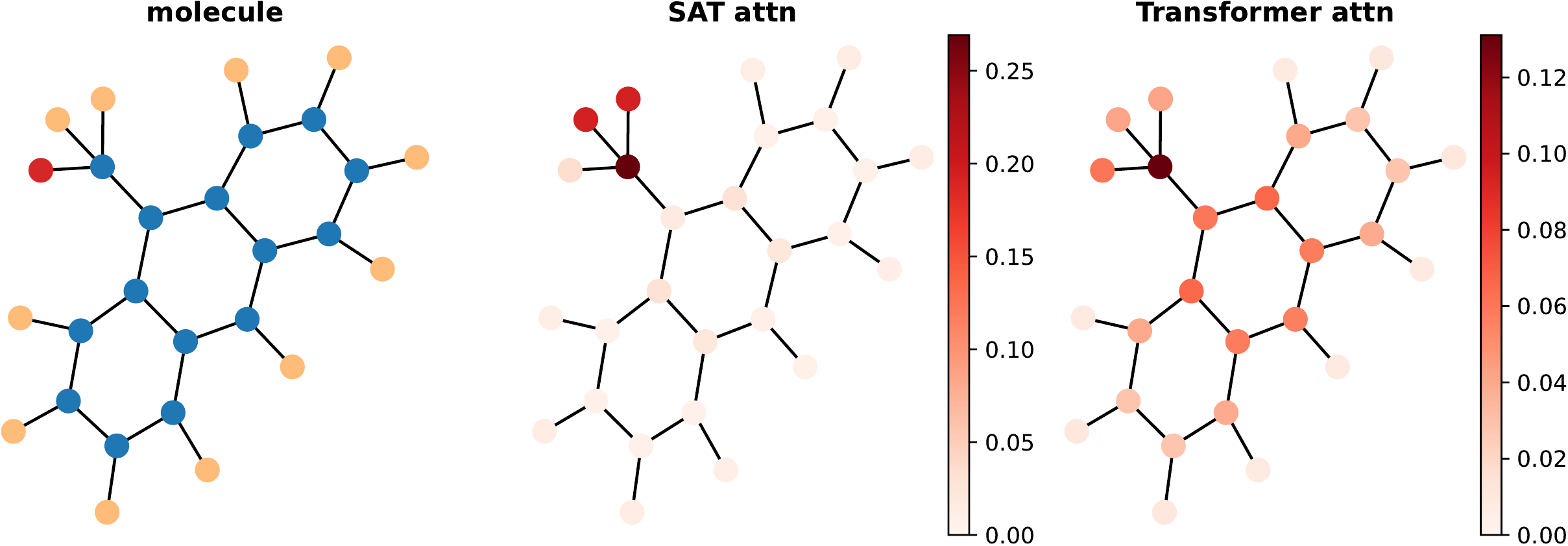}
    \includegraphics[width=.48\textwidth]{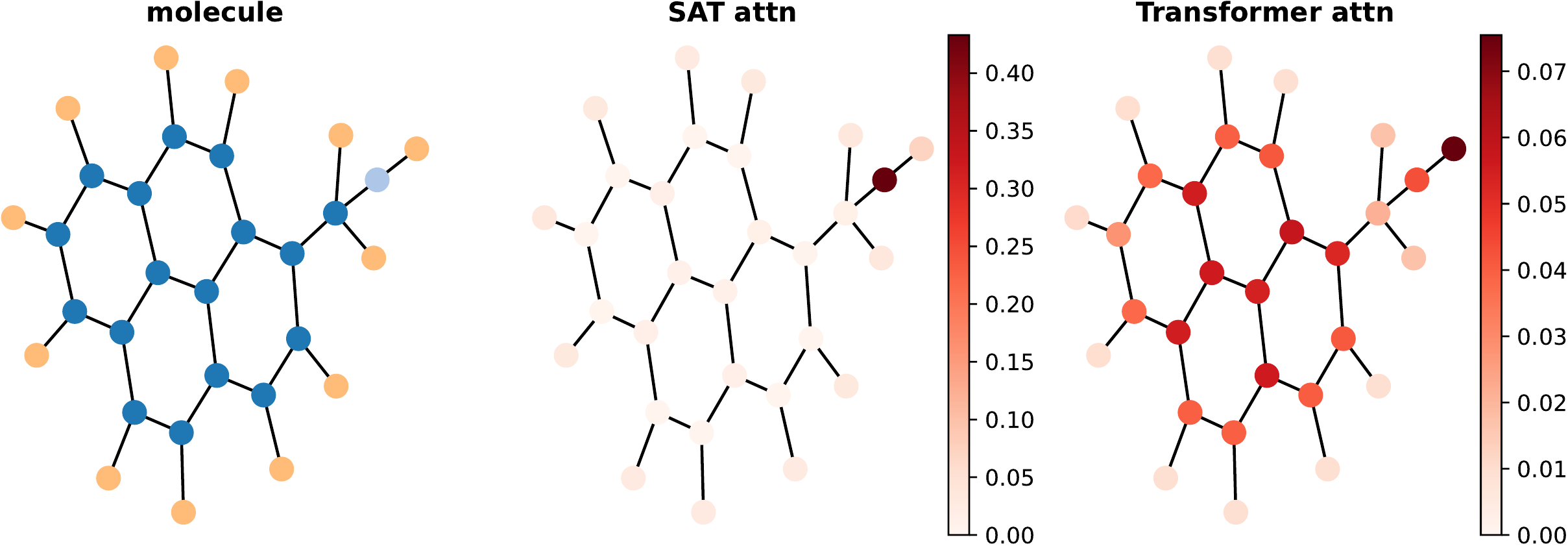}
    \caption{Attention visualization of SAT and the Transformer. The middle column shows the attention weights of the [CLS] node learned by our SAT model and the right column shows the attention weights learned by the classic Transformer with RWPE.}\label{fig:supp_attn_score}
    % }
    \end{center}

\section{Relationship to Subgraph Neural Networks and Graph Pooling}

\leo{
In the following we clarify the relationship (and differences) of SAT to Subgraph Neural Networks  \citep{alsentzer2020subgraph} as well as to the general topic of graph pooling.

\subsection{Differences to Subgraph Neural Networks}
Subgraph Neural Networks (SNN) \cite{alsentzer2020subgraph} explores explicitly incorporating position, neighborhood and structural information, for the purpose of solving the problem of subgraph prediction. SNN generates representations at the level of subgraph (rather than node). SAT, on the other hand, is instead motivated by modeling the \emph{structural interaction} (through the dot-product attention) between nodes in the Transformer architecture by generating node representations that are structure-aware. This structure-aware aspect is achieved via a structure extractor, which can be any function that extracts local structural information for a given node and does not necessarily need to explicitly extract subgraphs. For example, the $k$-subtree GNN extractor does not explicitly extract the subgraph, but rather only uses the node representation generated from a GNN. This aspect also makes the $k$-subtree SAT very scalable. In contrast to SNN, the input to the resulting SAT is not subgraphs but rather the original graph, and the structure-aware node representations are computed as the query and key for the dot-product attention at each layer.  

\subsection{Relationship to Graph Pooling}
In GNNs, structural information is traditionally incorporated into node embeddings via the neighborhood aggregation process. A supplemental way to incorporate structural information is through a process called \emph{local pooling}, which is typically based on graph clustering \citep{ying2018hierarchical}. Local pooling coarsens the adjacency matrix at layer $l$ in the network by, for example, applying a clustering algorithm to cluster nodes, and then replacing the adjacency matrix with the cluster assignment matrix, where all nodes within a cluster are connected by an edge. An alternative approach to the pooling layer is based on sampling nodes \citep{gao2019graph}. While \citet{rethinkpooling2020} found that these local pooling operations currently do not improve performance relative to more simple operations, local pooling could in theory be incorporated into the existing SAT's $k$-subtree and $k$-subgraph GNN extractors, as it is another layer in a GNN.
}

\begin{comment}
    "Graph pooling based on graph clustering can be used to identify structural similarity between nodes.
Understand the structural information of a graph using subgraphs or graph pooling."
\end{comment}

\end{figure}

\paragraph{Visualization of attention scores.}
Here, we provide additional visualization examples of attention scores of the [CLS] node from the Mutagenicity dataset, learned by SAT and a vanilla Transformer. Figure~\ref{fig:supp_attn_score} provides several examples of attention learned weights. SAT generally learns sparser and more informative weights even for very large graph as shown in the left panel of the middle row.

\end{document}